\documentclass[letterpaper]{article} 
\usepackage{aaai2026}  
\usepackage{times}  
\usepackage{helvet}  
\usepackage{courier}  
\usepackage[hyphens]{url}  
\usepackage{graphicx} 
\usepackage{natbib}  
\usepackage{caption} 
\usepackage{amsmath, amsfonts, amsthm, amssymb}
\usepackage{enumitem}
\usepackage[ruled, vlined, linesnumbered]{algorithm2e}
\usepackage{xcolor}
\usepackage{pifont}
\usepackage{subcaption}
\usepackage{multirow}
\usepackage{booktabs}

\DeclareMathOperator{\Tokenize}{\texttt{TKN}}
\DeclareMathOperator{\Adapter}{\texttt{ADP}}
\DeclareMathOperator{\Classifier}{\texttt{CLS}}
\DeclareMathOperator{\Retriever}{\texttt{RET}}
\DeclareMathOperator{\Generator}{\texttt{GEN}}
\DeclareMathOperator{\Transformers}{\texttt{TRF}}
\DeclareMathOperator{\topK}{\texttt{topK}}

\newtheorem{theorem}{Theorem}[section]
\newtheorem{lemma}[theorem]{Lemma}
\newtheorem{assumption}[theorem]{Assumption}
\newtheorem{proposition}[theorem]{Proposition}

\newcommand{\norm}[1]{\left\| #1 \right\|}
\newcommand{\dotproduct}[2]{\left\langle #1, #2 \right\rangle}
\newcommand{\maxi}[1]{\max_{i \in \mathcal{M}}{\left( #1 \right)}}
\newcommand{\mini}[1]{\min_{i \in \mathcal{M}}{\left( #1 \right)}}

\newcommand{\bbE}[1]{\mathbb{E}{\left[ #1 \right]}}
\newcommand{\bbEi}[1]{\mathbb{E}_{\xi_i \sim \mathcal{D}_i}{\left[ #1 \right]}}
\newcommand{\bbEit}[1]{\mathbb{E}_{\xi_i^{(t)} \sim \mathcal{D}_i}{\left[ #1 \right]}}
\newcommand{\bbEt}[1]{\mathbb{E}_{\substack{\{ \xi_i^{(t)} \sim \mathcal{D}_i \}_{i \in \mathcal{M}} \\ \{ \epsilon_i^{(t)} \sim \mathcal{N}_i^{(t)} \}_{i \in \mathcal{M}}} }{\left[ #1 \right]}}

\SetCommentSty{mycommfont}

\usepackage{newfloat}
\usepackage{listings}
\DeclareCaptionStyle{ruled}{labelfont=normalfont,labelsep=colon,strut=off} 
\usepackage{hyperref}

\title{Federated Learning with Ad-hoc Adapter Insertions: The Case of Soft-Embeddings for Training Classifier-as-Retriever}
\author{
    Marijan Fofonjka$^\dagger$,
    Shahryar Zehtabi$^{\dagger,}$\footnote{Work done during a summer internship at webAI in 2025.},
    Alireza Behtash$^\dagger$,
    Tyler Mauer$^\dagger$, and
    David Stout$^\dagger$
}
\affiliations{
    $^\dagger$webAI, $^*$Purdue University\\

    $^\dagger$\{marijan.fofonjka, shahryar.zehtabi, alireza.behtash, tyler, david\}@webai.com\\
    $^*$szehtabi@purdue.edu \\
}

\begin{document}

\maketitle

\begin{abstract}
When existing retrieval-augmented generation (RAG) solutions are intended to be used for new knowledge domains, it is necessary to update their encoders, which are taken to be pre-trained large language models (LLMs). However, fully fine-tuning these large models is compute- and memory-intensive, and even infeasible when deployed on resource-constrained edge devices. We propose a novel encoder architecture in this work that addresses this limitation by using a frozen small language model (SLM), which satisfies the memory constraints of edge devices, and inserting a small adapter network before the transformer blocks of the SLM. The trainable adapter takes the token embeddings of the new corpus and learns to produce enhanced soft embeddings for it, while requiring significantly less compute power to update than full fine-tuning. We further propose a novel retrieval mechanism by attaching a classifier head to the SLM encoder, which is trained to learn a similarity mapping of the input embeddings to their corresponding documents. Finally, to enable the online fine-tuning of both (i) the encoder soft embeddings and (ii) the classifier-as-retriever on edge devices, we adopt federated learning (FL) and differential privacy (DP) to achieve an efficient, privacy-preserving, and product-grade training solution. We conduct a theoretical analysis of our methodology, establishing convergence guarantees under mild assumptions on gradient variance when deployed for general smooth non-convex loss functions. Through extensive numerical experiments, we demonstrate (i) the efficacy of obtaining soft embeddings to enhance the encoder, (ii) training a classifier to improve the retriever, and (iii) the role of FL in achieving speedup.
\end{abstract}

\section{Introduction}

\begin{figure*}
    \centering
    \includegraphics[width=0.8\linewidth]{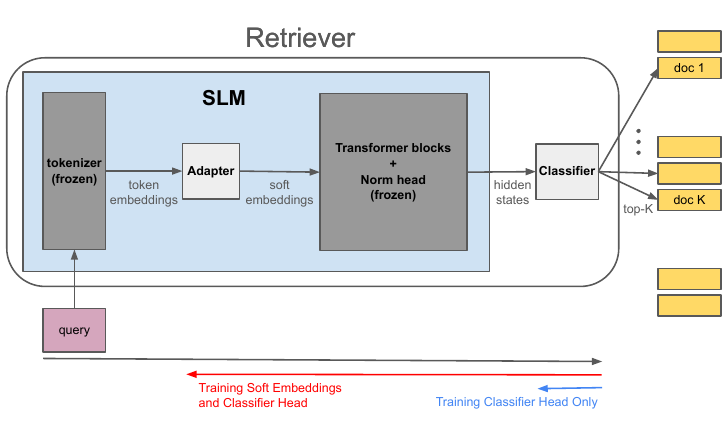}
    \caption{Overview of our CaR approach, where we make two orthogonal improvements to the retriever architecture. In contrast to existing RAG training algorithms, where the full retriever LLM is fine-tuned, we instead freeze it in our work and add a small Adapter network for PEFT. Second, instead of relying on maximum inner-product search for retrieval, we attach a Classifier head to the retriever to obtain similar documents to a query. Note that this architecture enables us to train the retriever in two ways, one where we only train the classifier head, and the other where we fine-tune both the soft embedding adapter network and the classifier head.}
    \label{fig:softembeddings}
\end{figure*}

Large language models (LLMs) have shown promising breakthroughs in recent years in a plethora of natural language processing applications \cite{radford2019language, devlin2019bert}. As these immensely complex models \cite{touvron2023llama} are pre-trained on vastly large datasets \cite{weber2024redpajama}, they exhibit high generalization across a lot of use cases due to the comprehensive knowledge they store in their parameters \cite{petroni2019language}. To further improve the performance of LLMs in various downstream tasks, adaptation techniques such as parameter fine-tuning have been adopted \cite{brown2020language}. Another approach has been to add non-parametric knowledge to LLMs through retrieval-augmented generation (RAG), where several similar documents to the input query are indexed from a given text corpus and augmented to the query as context \cite{lewis2020retrieval}. Although RAG has proven to be a prominent solution for knowledge-intensive tasks \cite{gao2023retrieval}, existing algorithms are limited in the following two aspects: (i) they either rely on the knowledge of the pre-trained LLM, not fine-tuning it at all for the specific target domain, or fully fine-tune the LLM encoder which can be compute-intensive; (ii) to find the most similar documents to an input query, they use heuristic similarity functions like the dot-product which are suboptimal \cite{khattab2020colbert}. To address both of the issues mentioned in (i), we propose a novel encoder architecture in this paper by inserting small trainable adapters to a frozen small language model (SLM), which improves the encoding mechanism by steering the document embeddings toward the domain of the new corpus. To further address (ii), we propose a trainable similarity function by attaching a classifier head to the SLM encoder, guiding the retriever towards training a more intelligent similarity calculator.

On the other hand, with the ever-increasing size of large models, both deploying and training them over memory-constrained devices at the edge remains a persistent challenge \cite{panchal2024thinking}. To mitigate the deployment issue, some works have focused on load offloading in which the LLM is decomposed and its blocks are distributed among multiple host servers \cite{borzunov2023petals}, and another line of research has studied quantization of LLMs \cite{behtash2025universality}. For cost-effective training of LLMs, parameter-efficient fine-tuning (PEFT) \cite{ding2023parameter} has been proposed to partially fine-tune the LLMs and circumvent complete model training, and federated learning (FL) has received attention as a promising solution for collaboratively optimization among multiple clients \cite{sani2024photon}. Nevertheless, a memory-efficient and cost-effective retriever architecture for RAG, which also needs to support distributed load offloading, is unexplored in the literature. Our proposed approach resolves this issue by inserting small neural layers to a frozen memory-confined SLM, that enable it to obtain enhanced soft embeddings of a new corpus while supporting distributed operation.

Furthermore, data residing at the edge are often private, requiring privacy-preserving training strategies. While FL provides an inherent layer of privacy by communicating model parameters rather than raw data \cite{kairouz2021advances}, the privacy of clients can be further reinforced by employing differential privacy (DP) techniques \cite{li2022does}. DP has been investigated for language modeling by carefully clipping the gradients and injecting noise into the model parameters, preventing any information leakage from individuals to the potentially untrustworthy aggregator \cite{li2022large}. Motivated by the lack of a differentially private FL approach for RAG training, we present a theoretically-grounded hybrid fixed/adaptive DP method in this work which is applied to the adapter and the classifier weights, enabling the clients to achieve local DP.

\textbf{Motivation.} RAG methods use a pre-trained LLM to obtain embeddings of new corpora, which are then employed to retrieve related documents to an input query using similarity functions. However, existing RAG architectures are (i) not suitable for distributed training on memory-constrained edge devices, (ii) do not efficiently adapt their LLM encoders to compute updated embeddings of the new corpus, and (iii) utilize suboptimal similarity functions like maximum inner-product search.
To address these issues, as illustrated in Fig.~\ref{fig:softembeddings}, (i) we first adopt a frozen pre-trained small language model (SLM) to satisfy the memory constraints of edge devices, and use FL for efficient distributed training among multiple edge devices. We also introduce DP to further preserve the privacy of participating clients; (ii) we then introduce the notion of \textit{soft embeddings} to enable the SLM encoder to adapt to a new corpus, by inserting smaller trainable adapters before the transformer blocks of the SLM; (iii) we finally propose a novel \textit{Classifier-as-Retriever (CaR)} methodology, where the similarity of documents to an input query in a QA corpus is determined by the top-$K$ outputs of the classifier. In this paper, our aim is thus to answer the following key question:
\begin{itemize}[leftmargin=*]
    \item[] \textit{How can we design an enhanced classifier-as-retriever approach which steers the encoder embeddings towards a new corpus, while satisfying resource constraints in a federated training process and preserving differential privacy?}
\end{itemize}

\textbf{Contributions.} 
We answer this question by making the following novel contributions.
\begin{itemize}
    \item We design a novel language encoder architecture with memory and communication considerations in mind. Since it is extremely costly to fully fine-tune an LLM in a FL setup, we instead use a \textit{frozen SLM} and \textit{insert smaller trainable adapters after the token embeddings and before the transformer blocks}. These adapters, which are modeled as square transformation matrices, learn \textit{soft embeddings} of the new corpus that was unseen during the pre-training of the SLM encoder.

    \item In contrast to conventional similarity search methods, we propose a novel retrieval algorithm by \textit{attaching a classifier head to the SLM encoder}. Enabled by the classifier, which maps the input query embeddings to their corresponding documents for a QA dataset, we achieve a mechanism to steer the soft embeddings to the domain of the new corpus. Through the process of training the full retrieval model consisting of the frozen pre-trained SLM, soft embedding adapters, and the classifier head on a task-specific QA corpus, we obtain a domain-specific and fast retrieval model. During inference, the top-$K$ output documents of the classifier for a given input prompt are then utilized for RAG.

    \item We adopt FL for online fine-tuning of our proposed RAG methodology using local client queries. To achieve communication efficiency for FL, we freeze the SLM encoder and only train the soft embedding adapters and classifier heads. To further enhance the privacy of clients, we employ either a fixed or adaptive DP technique, which clips local gradients of the clients and adds Gaussian noise to their model parameters. We further conduct rigorous convergence analysis of our methodology for general smooth non-convex loss functions and under mild assumptions on gradient variances, for both fixed and adaptive DP strategies.

    \item Through numerical experiments, we illustrate that our retrieval approach based on soft embedding adapters and classifier head culminates in a superior performance compared to conventional search methods for retrieval. Furthermore, we demonstrate that our FL distributed training methodology achieves significant speedup to their centralized counterparts, while still matching them in classification accuracy. Finally, we show that adding DP to our FL methodology ensured the privacy of clients while only resulting in a small drop in the achieved performance.
\end{itemize}

\section{Related Work}

\subsection{Fine-Tuning LLMs}

Parameter-efficient fine-tuning (PEFT) has attracted significant attention in recent years as an alternative to full model fine-tuning \cite{he2022towards}. For example, adapter tuning \cite{houlsby2019parameter} fine-tunes only small-sized adapters inserted into the LLM, while prefix tuning \cite{li2021prefix} and prompt tuning \cite{lester2021power} prepend small and fixed prefix tokens to the input and only fine-tune them. On the other hand, low-rank adaptation (LoRA) \cite{hu2022lora} has been proposed as an alternative that does not alter the original LLM architecture, but adds low-rank matrices to its constituent blocks to fine-tune the LLM. However, the addition of trainable LoRA modules to all transformer blocks of an LLM is not suitable when we aim to offload the load of the transformers across multiple servers in FL (see Sec.~\ref{ssec:offload}), as it increases memory and communication overhead at each participating node.

\subsection{Federated Learning and Differential Privacy}

LLM fine-tuning requires high-quality data, which can be difficult to obtain due to client privacy concerns \cite{zhang2024towards}. FL addresses this challenge by distributing the training process among data-holding clients, while still achieving performance comparable to centralized training \cite{kuang2024federatedscope}. Several studies have also explored the development of PEFT methods in FL settings—for example, \cite{wang2024flora, bai2024federated} proposed LoRA-based approaches tailored for FL.

To further enhance the privacy of clients participating in the FL process, DP techniques have been investigated \cite{sun2024improving, tran2025privacy}. In this work, we focus on local DP, which ensures client-side privacy—unlike central DP, which is applied by a central aggregator \cite{mcmahan2017learning}. The benefits of combining FL with DP to establish a two-tiered privacy framework are discussed in \cite{rodriguez2020federated}.

\subsection{Retrieval-Augmented Generation}

RAG is an elegant approach to incorporate retrieval-based knowledge into pre-trained LLMs, allowing them to provide references to their responses and reduce hallucinations \cite{huang2025survey}. The RAG pipeline is comprised of a retriever and a generator, with initial implementations using a maximum inner-product search (MIPS) to perform a similarity search \cite{lewis2020retrieval}. The authors of \cite{borgeaud2022improving} improve this by chunking the retrieval documents into token streams and then employing a frozen kNN retriever followed by trainable attention layers for an overall enhanced RAG. In this work, we present a novel retrieval approach called ``Classifier-as-Retriever" (CaR), in which we move away from conventional similarity search methods, to achieve an edge-deployable solution (see Sec.~\ref{sec:method} for more details). The addition of a trainable classifier, rather than relying on a static MIPS function, gives our approach the flexibility for complex similarity mapping.

\section{Methodology} \label{sec:method}

In this section, we describe the details of our proposed methodology. In Sec.~\ref{ssec:adapter}, we present our novel idea of adapter insertion to obtain soft embeddings of a new corpus. Then we describe our classifier-based retrieval approach for RAG in Sec.~\ref{ssec:classifier}. Finally, in Secs.~\ref{ssec:fl} and \ref{ssec:dp}, we outline the details of our federated learning and differential privacy solution, respectively.

\subsection{Adapter Insertion for Soft Embeddings} \label{ssec:adapter}

Since the encoders in RAG are usually taken to be LLMs that are highly parametrized and costly to communicate, it is inefficient to fully fine-tune them to obtain adapted embeddings for a new corpus. Motivated by this challenge, we substitute an SLM as the encoder and freeze its parameters. To fine-tune the embeddings, we instead insert a smaller adapter network into the SLM encoder and make it trainable, which helps us to learn soft embeddings of the new corpus. The adapter architecture is a square transformation matrix that sits between the token embeddings and the first transformer block.

Let the SLM (in the retriever block) parameters, that is, the tokenizer embedding layer and transformer blocks, be denoted as $\phi_e$ and $\phi_t$, respectively. Also, let the adapter parameters be denoted as $\theta_a$. With these definitions, for a given input query $x$, the token embeddings will be obtained as $\Tokenize_{\phi_e}(x)$. Subsequently, the soft embeddings are calculated as $\Adapter_{\theta_a}(\Tokenize_{\phi_e}(x))$, where $\Adapter$ is a transformation matrix parametrized by $\theta_a$. Next, the final hidden states of the SLM are obtained as $\Transformers_{\phi_t}(\Adapter_{\theta_a}(\Tokenize_{\phi_e}(x)))$ after passing through the transformer blocks $\Transformers$.

\subsection{Classifier-as-Retriever} \label{ssec:classifier}

In our work, we present a novel retrieval mechanism by attaching classifier heads to the retriever. Our CaR approach leverages the knowledge obtained in the soft embeddings to score the retrieval documents in terms of their similarity to the input query. The top-$K$ outputs of this classifier are then augmented to the input query to complete the retrieval phase of our RAG approach. Ultimately, the augmented inputs are fed into the LLM generator to create enhanced responses to the original client query.

Let the classifier parameters be defined as $\theta_c$, and the corresponding ground-truth answer to the query $x$ in the new corpus be denoted as document $y$. Consequently, the final output of the retriever will be computed as
\begin{equation} \label{eqn:retriever}
    \begin{aligned}
        \hat{y} & = \Retriever_{\phi_e, \theta_a, \phi_t, \theta_c}(x, K)
        \\
        & = \topK(\Classifier_{\theta_c}(\Transformers_{\phi_t}(\Adapter_{\theta_a}(\Tokenize_{\phi_e}(x)))), K),
    \end{aligned}
\end{equation}
in which the classifier function takes the transformer outputs which are operated on the soft embedding, and predicts the label of a given query. Then, the top-$K$ matched documents are returned for the retriever. The goal of our classifier training is to make accurate predictions $\hat{y}$ given inputs $x$, which are evaluated by the ground-truth output labels $y$.
Finally, letting $\phi_g$ denote the parameters of the LLM generator, the response of our RAG solution will be given as $\Generator_{\phi_g}(x, \Retriever_{\phi_e, \theta_a, \phi_t, \theta_c}(x, K))$, where the LLM generator take the original query alongside the top-$K$ retrieved documents $\hat{y}$ and produces a response. We would like to reiterate that while $\theta_a$ and $\theta_c$ are trainable parameters in our formulation, $\phi_e$, $\phi_t$ and $\phi_g$ are frozen during training.

\subsection{Federated Fine-Tuning} \label{ssec:fl}

We consider a distributed training setting with $m$ clients, collected in the set $\mathcal{M} = \{ 1, ..., m \}$. In our FL setup, each client $i \in \mathcal{M}$ trains only on its own portion of the training dataset with size $n_i$, denoted as $\mathcal{D}_i = \{ (x_{i,1}, y_{i,1}), ..., (x_{i,n_i}, y_{i,n_i}) \}$. Each client runs a gradient-based optimizer on its local dataset, and then it shares the trained parameters of the soft embedding layer (local adapter parameters $\theta_{a,i}$) and the local classifier head ($\theta_{c,i}$) with the central aggregator. We emphasize that both $\theta_{a,i}$ and $\theta_{c,i}$ parameters are significantly smaller compared to the SLM encoder parameters $\phi_e$ and $\phi_t$ and the LLM generator $\phi_g$, all of which are frozen on the clients during our federated fine-tuning process. We adopt the FedAvg \cite{mcmahan2017communication} method as the aggregation operation in our work, in which client updates are aggregated via weighted averaging to train the global model. Subsequently, the central aggregator sends the updated global model back to the clients to start the next round of training. Each client updates its local model for $E$ epochs before sharing it with the server, for a total of $T$ communication rounds.

Let $\theta = \{ \theta_a, \theta_c \}$ denote the set of all trainable parameters. The FL global and local objective functions can hence be formulated as
\begin{equation} \label{eqn:objectives}
    \begin{gathered}
        F(\theta) = \sum_{i=1}^m{\frac{n_i}{n} F_i(\theta)},
        \\
        \hspace{-5mm} F_i(\theta) \hspace{-0.5mm} = \hspace{-0.5mm} \frac1{n_i}  \hspace{-0.5mm} \sum_{j=1}^{n_i} \hspace{-0.5mm} \ell(\Classifier_{\theta_c}(\Transformers_{\phi_t}(\Adapter_{\theta_a}(\Tokenize_{\phi_e}(x_j)))), y_j), \hspace{-10mm}
    \end{gathered}
\end{equation}
respectively, where $\ell(\hat{y}, y)$ is a loss function taking the predicted outputs $\hat{y}$ and true labels $y$ and computing a classification loss, and $n = \sum_{i=1}^m{n_i}$ is the total size of the combined datasets of clients. Note how in this formulation
$\phi_e$ and $\phi_t$ are the SLM encoder parameters that are frozen. To solve this objective function, the clients run gradient descent on their local parameters $\theta_i$, while the server performs the operations $\theta = \sum_{i=1}^m{\frac{n_i}{n} \theta_i}$ to obtain globally updated parameters. We finally note that to ensure security and availability, model parameters could be encrypted and exchanged over blockchain, allowing multiple training agents to subscribe and join training groups \cite{nguyen2021federated}. These agents could also receive micropayments as an incentive to contribute to the training process \cite{zhan2020learning}.

\subsection{Local Differential Privacy} \label{ssec:dp}

Our focus in this work is on both fixed and adaptive local DP applied on the client side \cite{andrew2021differentially}. This approach protects sensitive private datasets and model parameters before they are sent to a potentially unreliable central aggregator. DP is enforced on the parameter updates after each training iteration, by first clipping the gradient updates and then adding noise to them. Therefore, privacy for each client can be controlled by using the clipping threshold and the noise variance, which are applied to the parameters of the classifier head and soft embedding layers.

First, each client computes the difference
between its updated local parameters and the global model as $\Delta \theta_i = \theta_i - \theta$. Then, each client calculates a scaling factor as $\min(1, C_i / \| \Delta \theta_i \|)$, where $C_i$ is the clipping threshold for client $i \in \mathcal{M}$. Note that the product of this scaling factor and $\Delta \theta_i$ ensures that the resulting clipped parameters have a maximum norm of $C_i$. Then, a Gaussian noise with variance $\sigma^2$ is added to the clipped difference to obtain the final differentially private distortion to be added to local parameters, which will be the final parameters to be transmitted to the central aggregator. These operations can be summarized as
\begin{equation} \label{eqn:dp}
    \theta_i = \theta_i + \min\left( 1, {C_i}/{\norm{\Delta \theta_i}} \right) \Delta \theta_i + \mathcal{N}(\mathbf{0}, \sigma_{0,i}^2 C_i^2 I ).
\end{equation}
We also incorporate adaptive DP in our work. Please see details in Appendix~\ref{appendix:adaptive}.

\section{Scalability Remarks for Future Directions}

In this section, we discuss the scalability of our approach in two scenarios: when using large models that cannot be deployed on edge devices (Sec.~\ref{ssec:offload}), and when working with datasets containing large query–answer pairs (Sec.~\ref{ssec:moe}). We leave the exploration of these directions for future work.

\subsection{Distributed Load Offloading} \label{ssec:offload}

During inference, the majority of resource usage and computations occur in the transformer blocks. To optimize this in our federated setup, transformer blocks can be hosted in distributed clusters with frozen weights, allowing multiple clients to reuse pre-loaded transformer blocks \cite{borzunov2023petals}. Since prompts originate on the client's machine, the token embedding and soft embedding layers can be hosted locally, enabling clients to train it before passing the embedding vector to the distributed transformer blocks. Once the information is propagated through the transformer blocks, the last hidden state is returned to the client. Finally, the client normalizes the hidden state using its local norm head, then performs the classification task and trains the local classifier head. This approach enables multiple clients to fine-tune their models, comprised of soft embedding adapters and classifier heads, in a distributed manner, while also leveraging a shared frozen transformer block infrastructure. Clients can thus store only their fine-tuned weights without needing to run the entire model.

\subsection{Mixture of Experts} \label{ssec:moe}

As we are employing a classifier to train a similarity mapping between the input query embeddings (features $x$) and their corresponding documents (labels $y$), the number of input-output pairs can grow arbitrarily large with the size of the dataset, resulting in an increase in the size of the classifier head. Note that conventional MIPS-based similarity search also has this issue, since increasing the number of retrieval documents demands a larger embedding database and also more computations for inner-product search. However, our proposed CaR approach, combined with the distribution of FL, allows us to mitigate this challenge by using a combination of clustering and a mixture-of-experts (MoE) solution to cut down the number of necessary computations in the transformer block. In other words, we first run a clustering algorithm on the embeddings of the documents (labels), aiming for a balance in the number of documents that fall into each cluster. Then, we design an MoE with the experts being classifier heads intended to solely classify the query-document pairs in a particular cluster. Note that if the dataset size is significantly large, we can opt for multi-level clustering and hierarchical MoEs to ensure that the classifier block satisfies the memory and compute constraints of edge devices.

Through the process of training the MoEs, the gating mechanism of the MoE directs the incoming input embeddings towards the best expert that was trained on a cluster similar to that input embedding. This way, we ensure that the training of a CaR is made feasible by moving away from a fully connected classifier head to a sparse MoE, while ensuring that the similarity score is still calculated fast and accurately by training multiple classifier experts.

\section{Convergence Analysis}

In this section, we first present the iterative update relations used in our methodology in Sec.~\ref{ssec:updates}, which serve as the foundation for our analysis. Next, we outline the assumptions underlying our theoretical analysis in Sec.~\ref{ssec:assump}. Finally, we present our main results in Sec.~\ref{ssec:main}. All supplementary lemmas, propositions, and proofs are provided in Appendix~\ref{appendix:analysis}.

\subsection{Iterate Relations} \label{ssec:updates}

First, we note that the FL local and global loss functions outlined in Eq.~\eqref{eqn:objectives} can be rewritten as follows for the purpose of analysis
\begin{equation} \label{eqn:objective_modified}
    \begin{gathered}
        F(\theta) = \frac1m \sum_{i=1}^m{F_i(\theta)},
        \\
        \hspace{-9mm} F_i(\theta) \hspace{-0.5mm} = \hspace{-0.5mm} \frac{m}{n} \hspace{-0.5mm} \sum_{j=1}^{n_i}{\hspace{-0.5mm} \ell(\Classifier_{\theta_c}(\Transformers_{\phi_t}(\Adapter_{\theta_a}(\Tokenize_{\phi_e}(x_j)))), y_j)}. \hspace{-10mm}
    \end{gathered}
\end{equation}

At every iteration $t$, each client $i \in \mathcal{M}$ uses a mini-batch from its local dataset, i.e., $\xi_i^{(t)} \sim \mathcal{D}_i$, to compute a stochastic gradient $g_i(\theta^{(t)}, \xi_i^{(t)})$ for its current model $\theta^{(t)}$. Then, it takes a gradient step to update its local parameters as $\theta_i^{(t+0.5)} = \theta^{(t)} - \alpha_i^{(t)} g_i(\theta^{(t)}, \xi_i^{(t)})$. To determine whether clipping is required, we then calculate the update difference as
\begin{equation} \label{eqn:delta}
    \Delta \theta_i^{(t)} = \theta_i^{(t+0.5)} - \theta^{(t)} = - \alpha_i^{(t)} g_i(\theta^{(t)}, \xi_i^{(t)}),
\end{equation}
which results in the following final single-iteration update for each client
\begin{equation} \label{eqn:localupdate}
    \begin{aligned}
        & \theta_i^{(t+1)} = \theta_i^{(t)} + \min{\left( 1, \frac{C_i^{(t)}}{\| \Delta \theta_i^{(t)} \|} \right)} \Delta \theta_i^{(t)} + \epsilon_i^{(t)}
        \\
        & = \theta_i^{(t)} - \min{\left( 1, \frac{C_i^{(t)}}{\| \Delta \theta_i^{(t)}\|} \right)} \alpha_i^{(t)} g_i(\theta^{(t)}, \xi_i^{(t)}) + \epsilon_i^{(t)},
    \end{aligned}
\end{equation}
where
\begin{equation} \label{eqn:noise}
    \epsilon_i^{(t)} \sim \mathcal{N}_i^{(t)} = \mathcal{N}\left( \mathbf{0}, ( \sigma_{0,i}^2 )^{(t)} I + ( \sigma_{1,i}^2 )^{(t)} \|\Delta \theta_i^{(t)}\|^2 I \right)
\end{equation}
is a zero-mean Gaussian noise being added to the model with variance $( \sigma_{0,i}^2 )^{(t)} + ( \sigma_{1,i}^2 )^{(t)} \|\Delta \theta_i^{(t)}\|^2$. The aggregator then averages the local models to calculate the updated global model as
\begin{equation}  \label{eqn:globalupdate}
    \begin{aligned}
        & \theta^{(t+1)} = \frac1m \sum_{i=1}^m{\theta_i^{(t+1)}}
        \\
        & \begin{aligned}
            = \theta^{(t)} & - \frac1m \sum_{i=1}^m{\min{\left( 1, \frac{C_i^{(t)}}{\| \Delta \theta_i^{(t)} \|} \right)} \alpha_i^{(t)} g_i(\theta^{(t)}, \xi_i^{(t)})}
            \\
            & + \frac1m\sum_{i=1}^m{\epsilon_i^{(t)}}.
        \end{aligned}
    \end{aligned}
\end{equation}

\subsection{Assumptions} \label{ssec:assump}

Note that $\theta$ indicates the adapter network and the classifier in our paper, which are taken to be a transformation matrix and a multi-layer perceptron (MLP), respectively. Thus, we make the following smoothness assumption which is widely used for the analysis of these models in the literature \cite{bottou2018optimization}.
\begin{assumption}[Smoothness] \label{assump:smooth}
    The local loss function for each client $F_i(\theta)$ is $L_i$-smooth, i.e., its gradient $\nabla{F}_i(\theta)$ is $L_i$-Lipschitz continuous. We have
    \begin{equation}
        \norm{\nabla{F}_i(\theta) - \nabla{F}_i(\theta')} \le L_i \norm{\theta - \theta'},
    \end{equation}
    where $\theta, \theta'$ are any two arbitrary model parameters, and $L_i$ is a positive real number for all $i \in \mathcal{M}$.
\end{assumption}

We next make two Assumptions on the intra-client stochastic gradient variance and inter-client gradient variance \cite{bottou2018optimization}. Note that these assumptions do not enforce a uniform bound on the gradient noise and diversity but instead put a less strict variance bound on them. Moreover, the bounds for both Assumptions are not fixed global constants for all $\theta$, and instead the bounds depend on the global gradient norm $\norm{\nabla{F}(\theta)}^2$.
\begin{assumption}[Stochastic gradient variance] \label{assump:var}
    Let $g_i(\theta, \xi_i)$ be an unbiased estimate of the local gradient of client $i \in \mathcal{M}$, i.e., $\bbEi{g_i(\theta, \xi_i)} = \nabla{F}_i(\theta)$, where $\xi_i \in \mathcal{D}_i$ indicates a local mini-batch. Then, the average of local variances of $g_i(\theta, \xi_i)$ among all clients is bounded as
    \begin{equation}
        \frac1m \hspace{-0.5mm} \sum_{i=1}^m{\bbEi{ \norm{g_i(\theta, \xi_i) \hspace{-0.5mm} - \hspace{-0.5mm} \nabla{F}_i(\theta)}^2 }} \hspace{-0.5mm} \le \hspace{-0.5mm} \bar{\rho}_0^2 + \bar{\rho}_1^2 \norm{\nabla{F}(\theta)}^2,
    \end{equation}
    where $\bar{\rho}_0$ and $\bar{\rho}_1$ are positive real numbers.
\end{assumption}

\begin{assumption}[Gradient diversity] \label{assump:graddiv}
    The variance of local gradients $\nabla{F}_i(\theta)$ is bounded as
    \begin{equation}
        \frac1m \sum_{i=1}^m{\norm{\nabla{F}_i(\theta) - \nabla{F}(\theta)}^2} \le \bar{\zeta}_0^2 + \bar{\zeta}_1^2 \norm{\nabla{F}(\theta)}^2,
    \end{equation}
    where $\bar{\zeta}_0$ and $\bar{\zeta}_1$ are positive real numbers, and $\nabla{F}(\theta) = \frac1m \sum_{i=1}^m{\nabla{F}_i(\theta)}$.
\end{assumption}

We finally make the following assumption on the stochastic gradients commonly used in DP papers to analyze clipping \cite{zhang2022understanding, cheng2022differentially, shi2023make}. We emphasize that we will use this assumption only to analyze the clipping step of our approach.
\begin{assumption}[Stochastic gradient uniform bound] \label{assump:bound}
    The stochastic local gradient $g_i(\theta, \xi_i)$ is bounded for all $\xi_i \sim \mathcal{D}_i$ as
    \begin{equation}
        \norm{g_i(\theta, \xi_i)} \le B_i
    \end{equation}
    where $B_i$ is a positive real number. We also let $B = \maxi{B_i}$.
\end{assumption}

\subsection{Main Result} \label{ssec:main}

Our main analytical result is provided in the next theorem. For brevity, we have postponed all supporting proofs to Appendix~\ref{appendix:analysis}.

\begin{theorem}[Stationarity point for constant learning rate and fixed DP] \label{theorem:fixed}
    Let Assumptions~\ref{assump:smooth}-\ref{assump:bound} holds. Also, let a constant learning rate $\alpha_i^{(t)} = \alpha$ be adopted for all clients $i \in \mathcal{M}$, and a fixed DP algorithm be used, i.e., a fixed clipping threshold $C_i^{(t)} = C$ and a fixed noise variance given as $(\sigma_{0,i}^2)^{(t)} = \sigma_0^2 C^2$ and $(\sigma_{1,i}^2)^{(t)} = 0$. Then, Proposition~\ref{prop:finalloss} implies that if the learning rate $\alpha$ satisfies
    \begin{equation}
        \begin{aligned}
            \alpha < \min\Bigg( & \sqrt{\frac{2C}{3 \bar{L} (\bar{\rho}_1^2 + \bar{\zeta}_1^2 + 1) B (1 + \sigma_1^2) \Gamma_1}},
            \\
            & \frac2{3 \bar{L} (\bar{\rho}_1^2 + \bar{\zeta}_1^2 + 1) (1 + \sigma_1^2) \Gamma_2} \Bigg),
        \end{aligned}
    \end{equation}
    we will have
    \begin{equation} \label{eqn:fixed}
         \begin{aligned}
             \frac1{T+1} & \sum_{t=0}^T{\norm{\nabla{F}(\theta^{(t)})}^2} \le \underbrace{\frac{F(\theta^{(0)}) - F^\star}{\omega(T+1)}}_{\text{FedAvg}}
             \\
             & + \underbrace{\frac{\bar{L} C^2 (1 + \sigma_0^2)}{2\omega}}_{\substack{\text{DP clipping and} \\ \text{noise effect}}} + \underbrace{\frac{3 \bar{L} (\bar{\rho}_0^2 + \bar{\zeta}_0^2) \alpha^2}{2 \omega}}_{\substack{\text{Gradient variance} \\ \text{effect}}}.
         \end{aligned}
    \end{equation}
    in which $\omega = \min{( \frac{C}{B} ( 1 - \frac1{\Gamma_1} ), \alpha ( 1 - \frac1{\Gamma_2} ) )}$.

    \begin{proof}
        Follows from substituting the constant values given in the statement of the Theorem in Proposition~\ref{prop:finalloss}, and then reordering the inequality.
    \end{proof}
\end{theorem}

\textbf{Discussion.} Theorem~\ref{theorem:fixed} shows that for a constant learning rate and fixed DP, our FL approach achieves the same convergence rate $\mathcal{O}(1/T)$ as centralized training to a stationarity neighborhood of $\mathcal{O}(\frac{\bar{L} C^2 (1 + \sigma_0^2)}{2\omega} + \frac{3 \bar{L} (\bar{\rho}_0^2 + \bar{\zeta}_0^2) \alpha^2}{2 \omega})$. If the constant learning rate $\alpha$ and the fixed clipping threshold $C$ are chosen as $\alpha \propto 1 / \sqrt{T}$ and $C \propto 1 / \sqrt{T}$, then our approach achieves a slower convergence rate of $\mathcal{O}(1 / \sqrt{T})$, but the stationarity gap becomes zero. Moreover, note that our theoretical result in Theorem~\ref{theorem:fixed} can recover the results of existing methods which do not perform gradient clipping, by letting $C > \alpha B$, which corresponds to a clipping threshold large enough that will never be achieved by any of the clients.

We also conduct convergence analysis of our methodology when adaptive DP is employed, i.e., with an adaptive clipping threshold $C_i^{(t)}$ for all clients, and an adaptive additive noise which is based on $C_i^{(t)}$. See Appendix~\ref{appendix:adaptive} for all the details.

\section{Experimental Results}

\subsection{Setup} \label{ssec:setup}

\textbf{Dataset and Model.} We benchmark our methodology using three datasets, (i) SMS spam collection \cite{sms_spam_collection_228} with $5,574$ entries for binary classification, (ii) AG news \cite{del2005ranking} where 126,700 news topics are categorized to 4 classes, and (iii) arXiv metadata \cite{arxiv_org_submitters_2024} where 345,657 paper abstracts are categorized into 173 subject categories, which are all text classification datasets with labeled input-output pairs. As the SLM encoder, we employ the two models TinyLlama-1.1B-Chat-v1.0 \cite{zhang2024tinyllama} and Llama-3.2-3B-Instruct \cite{grattafiori2024llama} for different experiments, in which the token embedding layer is followed by 22 and 28 transformer blocks, respectively, and a final norm head layer. We insert a square transformation matrix as the adapter before the first transformer block, and attach a classifier head after the norm head. The classifier head consists of a fully-connected layer, with optional addition of a linear pre-classifier with a dropout layer between the layers. We take the hidden state to be the EOS token embeddings at the last layer unless otherwise stated. We use a batch size of $4$ across all of our experiments.

\textbf{Baselines.} We conduct two sets of experiments in our work: (i) we first train only the classifier head to test its effectiveness in doing similarity mapping, and (ii) then train both the adapter and the classifier head to investigate the role of adapter training in obtaining domain-specific soft embeddings in the LLM encoder. However, when training both the soft embeddings and the classifier head, doing backpropagation can become time-consuming since the small trainable adapter network resides at the beginning of the LLM, i.e., the frozen weights of the full LLM model is still in the computation graph. Therefore, we use a smaller model and a smaller dataset to illustrate the advantage of soft embeddings in Sec.~\ref{ssec:exp:soft}, and then focus on training a classifier only for the rest of the experiments.

\textbf{FL and DP.} In the FL approach, the dataset is split evenly between two clients. After each iteration, the clients exchange their locally trained weights with a central aggregator to update the global model. Training for the centralized implementation was conducted over 10 epochs, and 5 rounds of 2 epochs each for FL. To simulate DP, we apply adaptive clipping and adaptive noise to the classifier head training from the first round, using a noise scale parameter of $0.1$. When training the soft embeddings and classifier head jointly, we use fixed noise clipping with a value of $1.0$ to simulate a fixed-noise scenario. The soft embeddings are initialized with the identity matrix. This setup leads to faster convergence, as observed from the loss and performance metrics.

\textbf{Training Resources.} Our experiments are run using a Mac Studio equipped with an Apple M2 Ultra chip, featuring a 24-core CPU and 192 GB of unified memory. While our FL results are mainly in a simulated environment, we also conduct additional experiments for FL where the clients reside in different machines anc communicate over the gRPC protocol. In this setup, a cluster of four M4 Mac Mini machines with 64 GB RAM each was used, which are interconnected via Thunderbolt Bridge. One machine acts as the central aggregator, while the rest are the client machines.

\subsection{Comparison with MIPS} \label{ssec:exp:mips}

\begin{figure}[t]
    \centering
    \includegraphics[width=\linewidth]{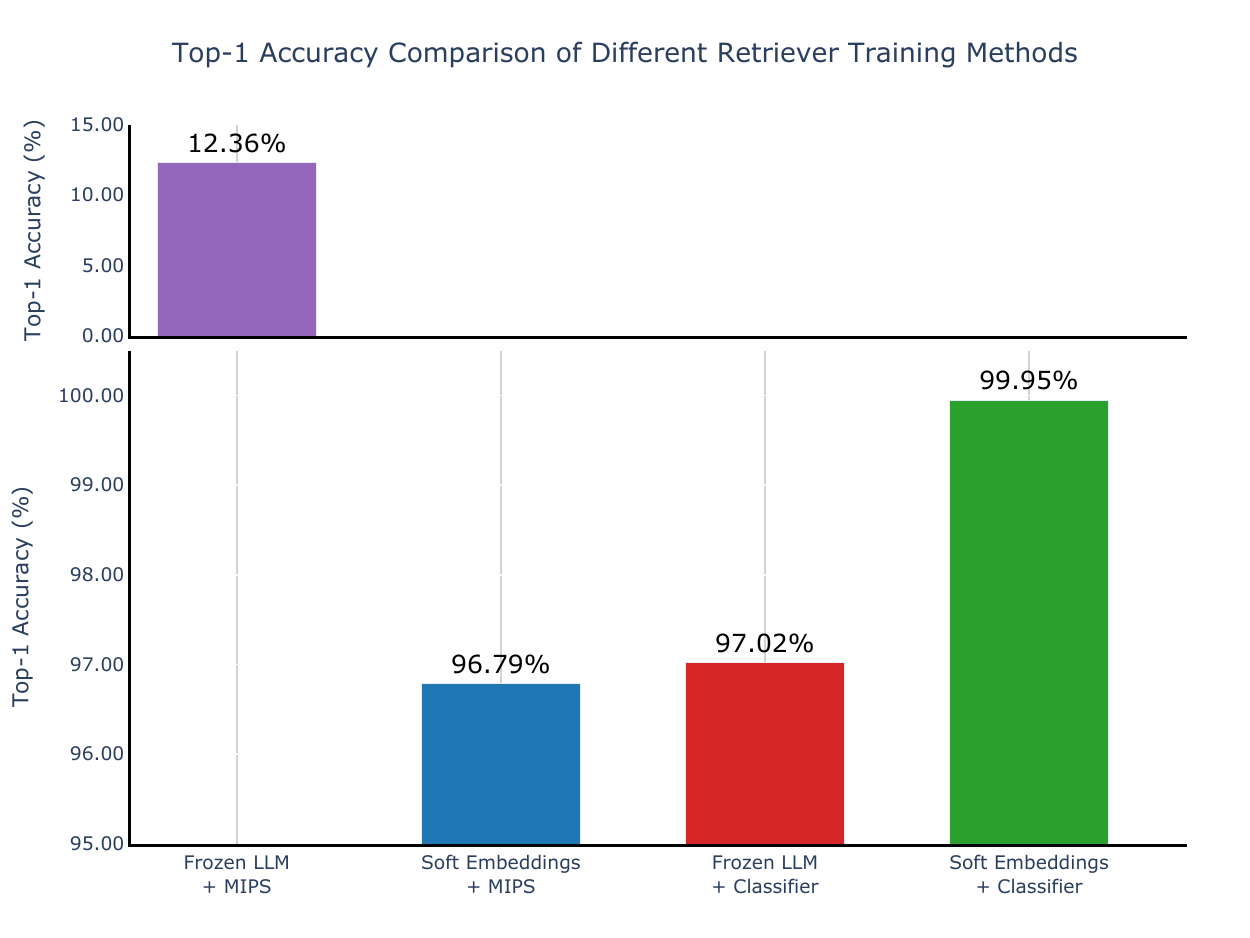}
    \caption{Comparison of maximum inner-product search (MIPS), which is the conventional retrieval method in existing RAG literature, with our proposed Classifier-as-Retriever methodology.}
    \label{fig:mips}
\end{figure}

In Fig.~\ref{fig:mips}, we compare our proposed soft embedding and classifier head ideas with conventional MIPS. For the purpose of comparison, we use TinyLlama on the SMS spam classification dataset. Our results show that when a pretrained LLM (like TinyLlama) is used on a specific downstream task that it has not seen, its retrieval performance is significantly low using MIPS (12.36\%). We thus conclude that some sort of training is required to increase the retrieval quality.

For this purpose, we insert a small trainable adapter, which is in the form of a transformation matrix, to the encoder LLM (TinyLlama) before its transformer blocks. We keep the similarity mapping function as MIPS in this experiment. Fig.~\ref{fig:mips} shows how our soft embeddings approach significantly improves the retrieval quality to 96.79\%. Training the adapter essentially helps transform the embeddings of the LLM to the soft embeddings of the new corpus, which in turn makes MIPS a good similarity mapping function.

Noting that training the soft embeddings still requires backpropagation of gradients across the full LLM, although its weights are frozen, we investigate a different training approach to increase the retrieval performance. Instead of attaching an adapter to the beginning of the LLM, we attach only a classifier head to the end of the frozen LLM and train it to map the hidden states to the corresponding labels in the dataset. Fig.~\ref{fig:mips} illustrates how this approach achieves a similar 97.02\% accuracy for retrieval. The classifier head is a light-weight and fast alternative for MIPS, which is faster to train and achieves better performance compared to it.

Finally, we combine the two components of our approach, namely the soft embeddings and the classifier head, to investigate the benefits of them combined together. Fig.~\ref{fig:mips} shows how this approach brings up the accuracy to 99.95\%, outperforming all previous baselines.

\subsection{Adapter Training for Soft Embeddings} \label{ssec:exp:soft}

\begin{figure*}[t]
    \centering
    \begin{subfigure}[t]{0.49\textwidth}
        \centering
        \includegraphics[width=\linewidth]{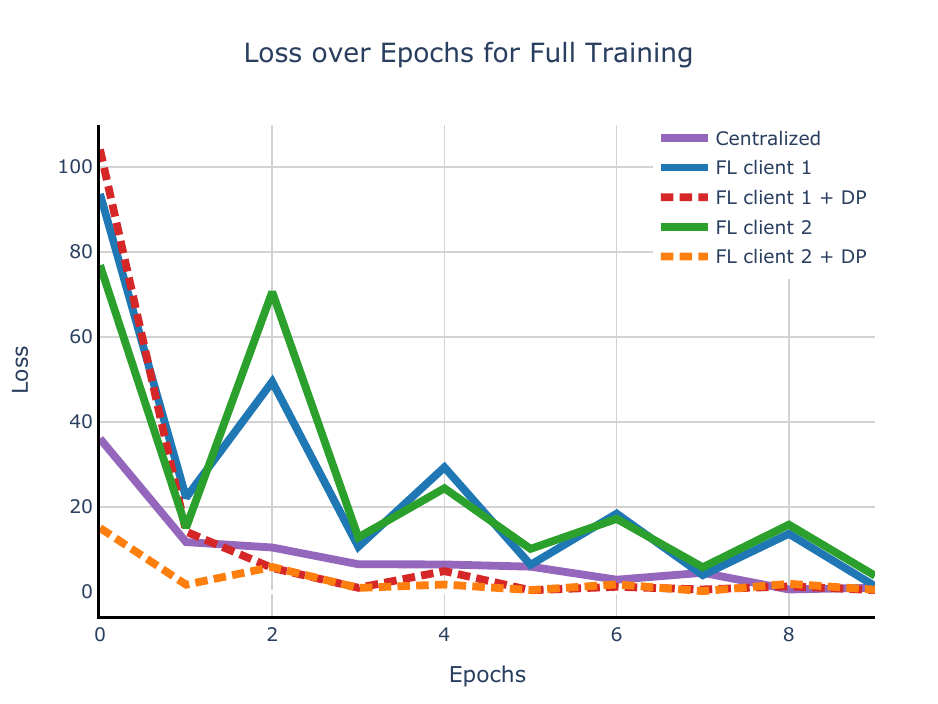}
        \caption{Loss for training soft embedding and classifier head.}
    \end{subfigure}
    \hfill
    \begin{subfigure}[t]{0.49\textwidth}
        \centering
        \includegraphics[width=\linewidth]{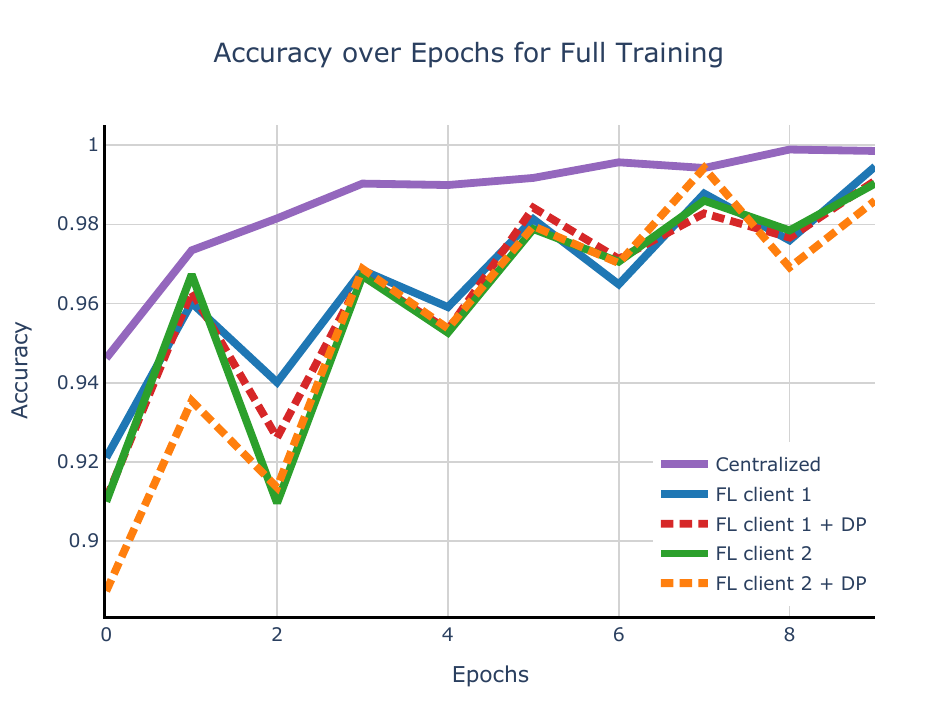}
        \caption{Accuracy for training soft embedding and classifier head.}
    \end{subfigure}
    \begin{subfigure}[t]{0.49\textwidth}
        \centering
        \includegraphics[width=\linewidth]{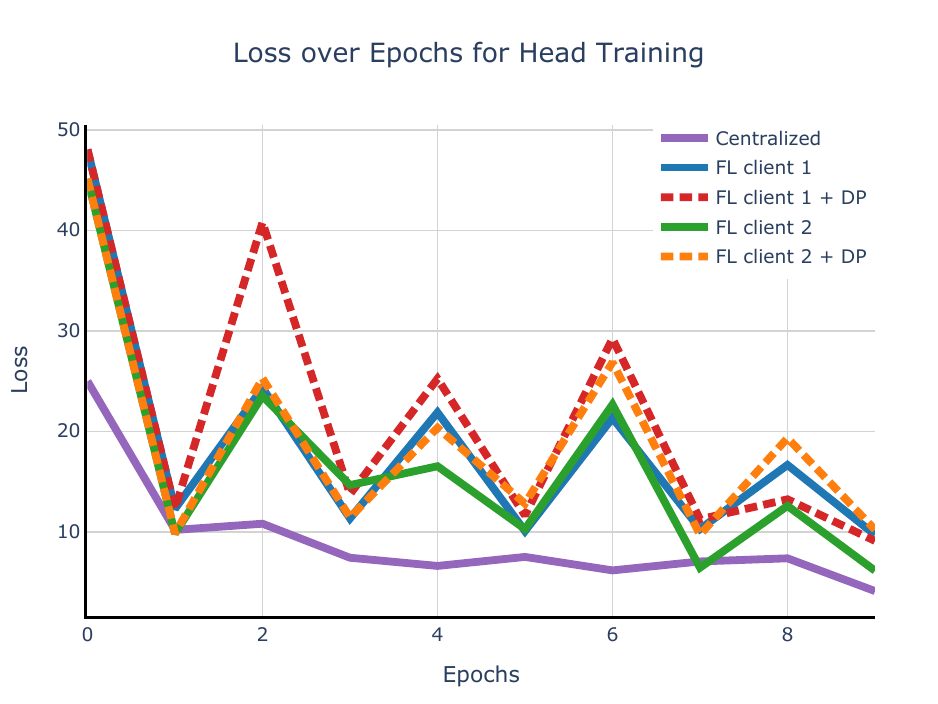}
        \caption{Loss for training classifier head only.}
    \end{subfigure}
    \hfill
    \begin{subfigure}[t]{0.49\textwidth}
        \centering
        \includegraphics[width=\linewidth]{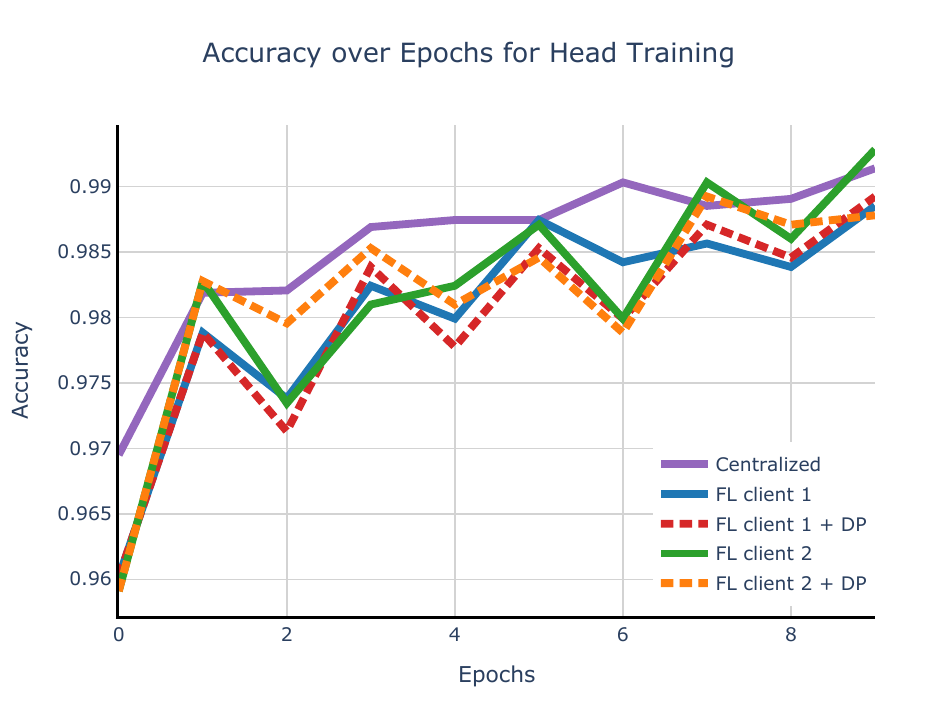}
        \caption{Accuracy for training classifier head only.}
    \end{subfigure}
    \caption{Training results for the SMS spam classification dataset using the TinyLlama model.}
    \label{fig:spam}
\end{figure*}

\begin{table}[t]
    {\small
    \setlength\tabcolsep{1.5pt}
    \centering
    \begin{tabular}{c|c|c|c|c}
        \toprule
        \multirow{3}{*}{Architecture} & \multirow{3}{*}{Scenario} & Top-$1$ & Total train & Distributed
        \\
        & & accuracy & \& val time & time
        \\
        & & (\%) & (min) & (min)
        \\
        \midrule
        \multirow{4}{*}{\begin{tabular}{c} Classifier head \\ only \end{tabular}} & Centralized & 97.02 & 0.20 & $\sim$ 0.10 \\
        \cline{2-5}
        & FL & 97.24 & 0.23 & $\sim$ 0.12 \\
        \cline{2-5}
        & \begin{tabular}{c} FL + DP\\(adaptive) \end{tabular} & 91.96 & 0.23 & $\sim$ 0.12 \\
        \midrule
        \multirow{4}{*}{\begin{tabular}{c} Soft embedding \\ adapter and \\ classifier head \end{tabular}} & Centralized & 99.95 & 32.14 & $\sim$ 32.14 \\
        \cline{2-5}
        & FL & 99.50 & 35.66 & $\sim$ 18 \\
        \cline{2-5}
        & \begin{tabular}{c} FL + DP\\(fixed) \end{tabular} & 99.64 & 35.85 & $\sim$ 18 \\
        \bottomrule
    \end{tabular}
    \caption{Summary of results for the global model, highlighting speed comparison of different methods.}
    \label{tab:spam}
    }
\end{table}

As training both the soft embeddings and the classifier head requires backpropagation through the transformer blocks, as shown in Fig.~\ref{fig:softembeddings}, it is still a relatively time-consuming process using our hardware (see Sec.~\ref{ssec:setup}). Therefore, in this initial section on training the soft embeddings, we present results from the relatively small SMS spam dataset using the TinyLlama model. We train the centralized model for 10 epochs, and the FL models for 5 rounds of 2 local epochs each for a fair comparison.

Our first observation is regarding the accuracy vs. speed trade-off of our FL approach. As seen in Table~\ref{tab:spam}, our distributed models achieve over 99\% accuracy, matching the performance of the centralized baseline, showing that FL maintains model performance while improving scalability and achieving a faster training time. In addition, Fig.~\ref{fig:spam} illustrates the FL client training process, and we see that they converge to the same centralized training performance in only $10$ epochs of training.

Moreover, training with soft embeddings and the classifier head simultaneously improves the overall performance compared to training with just the classifier head. As seen in Table~\ref{tab:spam}, there is a clear improvement in accuracy ($\sim 99\%$ vs. $97\%$) when enabling soft embeddings. This improvement is modest in this use case with the SMS Spam classification dataset since accuracy is already over $97\%$, but shows that the model better fits the data when training soft embeddings.

To investigate the effect of differential privacy (DP), we apply adaptive DP to the classifier head training. Our results in Fig.~\ref{fig:spam} and Table~\ref{tab:spam} shows that using DP results in only a slight drop in accuracy, while preserving client privacy. This demonstrates that DP can be effectively integrated with minimal performance loss.

\subsection{Distributed Training using gRPC} \label{ssec:exp:grpc}

\begin{figure*}[t]
    \centering
    \begin{subfigure}[t]{0.49\textwidth}
        \centering
        \includegraphics[width=\linewidth]{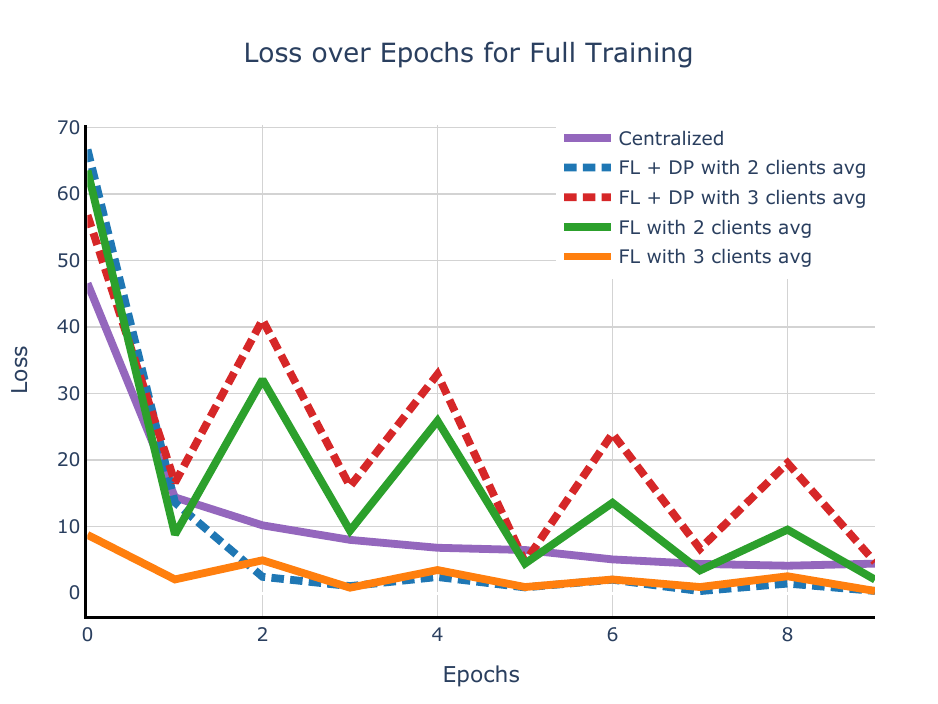}
        \caption{Loss for training soft embedding and classifier head.}
    \end{subfigure}
    \hfill
    \begin{subfigure}[t]{0.49\textwidth}
        \centering
        \includegraphics[width=\linewidth]{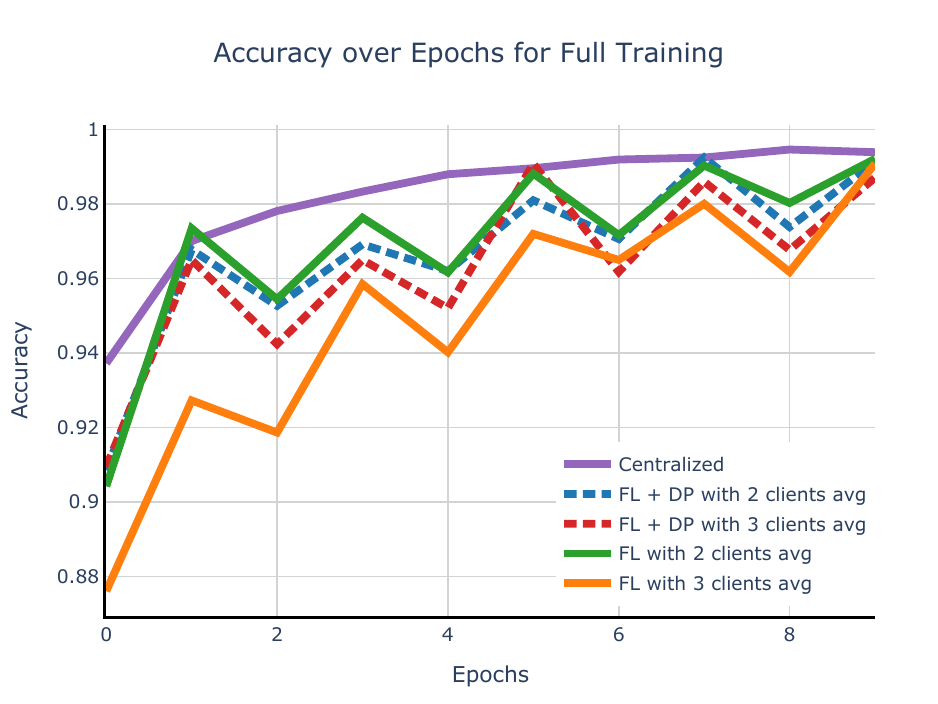}
        \caption{Accuracy for training soft embedding and classifier head.}
    \end{subfigure}
    \begin{subfigure}[t]{0.49\textwidth}
        \centering
        \includegraphics[width=\linewidth]{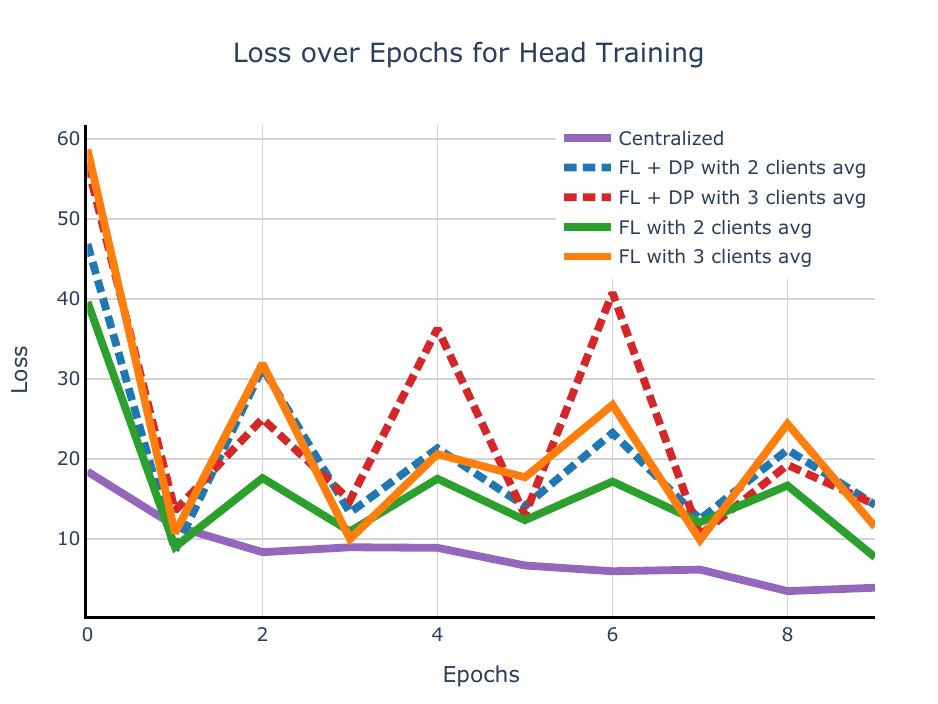}
        \caption{Loss for training classifier head only.}
    \end{subfigure}
    \hfill
    \begin{subfigure}[t]{0.49\textwidth}
        \centering
        \includegraphics[width=\linewidth]{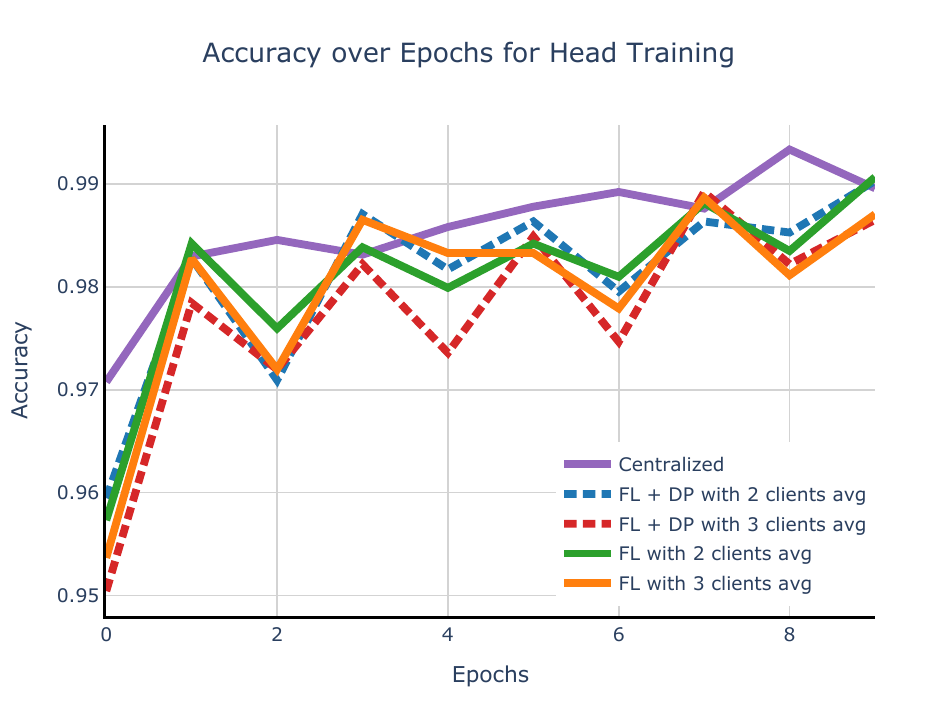}
        \caption{Accuracy for training classifier head only.}
    \end{subfigure}
    \caption{Training results for the SMS spam classification dataset with the TinyLlama model and using the gRPC protocol.}
    \label{fig:grpc}
\end{figure*}

\begin{table}[t]
    {\small
    \setlength\tabcolsep{1.5pt}
    \centering
    \begin{tabular}{c|c|c|c|c|c}
        \toprule
        \multirow{3}{*}{Architecture} & \multirow{3}{*}{Clients} & \multirow{3}{*}{Scenario} & Top-1 & Total train & \multirow{3}{*}{Speedup} \\
        & & & accuracy & \& val time & \\
        & & & (\%) & (min) & \\
        \midrule
        \multirow{7}{*}{\begin{tabular}{c} Classifier head \\ only \end{tabular}} & 1 & Centralized & 99.64 & 0.155 & - \\
        \cline{2-6}
        & \multirow{2}{*}{2} & FL & 99.28 & 0.139 & 1.11x \\
        \cline{3-6}
        & & \begin{tabular}{c} FL + DP\\(adaptive) \end{tabular} & 99.19 & 0.15 & 1.03x \\
        \cline{2-6}
        & \multirow{2}{*}{3} & FL & 99.37 & 0.139 & 1.11x \\
        \cline{3-6}
        & & \begin{tabular}{c} FL + DP\\(adaptive) \end{tabular} & 98.80 & 0.14 & 1.07x \\
        \midrule
        \multirow{7}{*}{\begin{tabular}{c} Soft embedding\\adapter and\\classifier head \end{tabular}} & 1 & Centralized & 99.80 & 55.40 & - \\
        \cline{2-6}
        & \multirow{2}{*}{2} & FL & 99.87 & 31.57 & \textbf{1.75x} \\
        \cline{3-6}
        & & \begin{tabular}{c} FL + DP\\(fixed) \end{tabular} & 99.55 & 32.92 & 1.68x \\
        \cline{2-6}
        & \multirow{2}{*}{3} & FL & 99.25 & 21.11 & \textbf{2.62x} \\
        \cline{3-6}
        & & \begin{tabular}{c} FL + DP\\(fixed) \end{tabular} & 99.52 & 22.09 & 2.51x \\
        \bottomrule
    \end{tabular}
    \caption{Summary of results for FL with the gRPC protocol, highlighting the achieved speedup.}
    \label{tab:grpc}
    }
\end{table}

The primary goal of this benchmarking process is to compare training speed and validate the correctness of the designed distribution protocol, specifically in terms of entity synchronization and model updates. Additionally, we aim to assess whether all approaches could successfully learn a simple classification task, such as spam detection. Similar to Sec.~\ref{ssec:exp:soft}, we run experiments using the TinyLlama model on the SMS spam dataset, with 10 epochs for centralized training and 5 rounds of 2 local epochs each for FL training. As seen in Fig.~\ref{fig:grpc}, our results demonstrate that the gRPC-based protocol functions correctly for distributed training: model accuracy consistently exceeds 99\% and improves with each iteration, while overall training loss steadily decreases.

Also as illustrated in Table~\ref{tab:grpc}, a key insight emerges when training more computationally intensive tasks, such as learning soft embeddings with a classifier head, where gradients must propagate back through a large language model (LLM). In such cases, where training time significantly exceeds network synchronization time, distributed federated learning yields substantial speedups. For instance, using two machines (each training on half of the dataset), we observe a 1.75x speedup. With three machines (each training on a third of the dataset), the speedup increases to 2.62x (21.11 minutes vs. 55.40 minutes). This demonstrates that distributing the workload across multiple machines enables significantly faster convergence of the global model. As expected, the speedup is not linear. It follows a trend more aligned with Amdahl’s or Gustafson’s Law.

Another observation we make from Table~\ref{tab:grpc} is that when training only the classifier head, the overall training time does not benefit significantly from distribution. This is because the hidden states are precomputed before the training begins, and prompts do not need to pass through the LLM during each iteration to calculate the loss. Since the entire LLM is frozen, hidden states are extracted just once, and training involves only updating the final classification layer. This setup makes classifier-head-only training an excellent example of how a large, shared LLM can be leveraged in a distributed environment. Multiple clients can independently use the shared LLM to extract hidden states and then perform lightweight, local training of their classifier heads.

\subsection{Results on Larger Models and Datasets}

Having established the superiority of our proposed soft embedding training in Secs.~\ref{ssec:exp:mips}, \ref{ssec:exp:soft}, and \ref{ssec:exp:grpc}, we turn to training only the classifier head in this section when evaluating our methodology on larger models and datasets.  Our goal is to demonstrate that our proposed method enables hosting the full LLM across multiple machines in a remote distributed cluster, which allows client machines, such as simple MAC notebooks, to benefit from large-scale distributed models for inference. Consequently, client machines only need to train lightweight, local classifier heads. Since we only need to extract the hidden state for each prompt (without backpropagation or gradient tracking), the remote cluster can simultaneously serve inference requests from multiple clients. These clients can then use the returned hidden states to train small classifier heads locally, which is typically just one or two simple linear layers. This approach also enables experimentation with different configurations, e.g. enabling a pre-classifier layer .

We will use more realistic datasets in this section, i.e., AG News and arXiv metadata, to show the practical utility of our approach. We also explore the impact of larger models as they typically provide higher-quality embeddings and better attention mechanisms, and thus, using their hidden states leads to better performance. We use a learning rate of $10^{-5}$ for AG news, and $10^{-3}$ for arXiv metadata.

\subsubsection{AG news}

\begin{figure}[t]
    \centering
    \includegraphics[width=\linewidth]{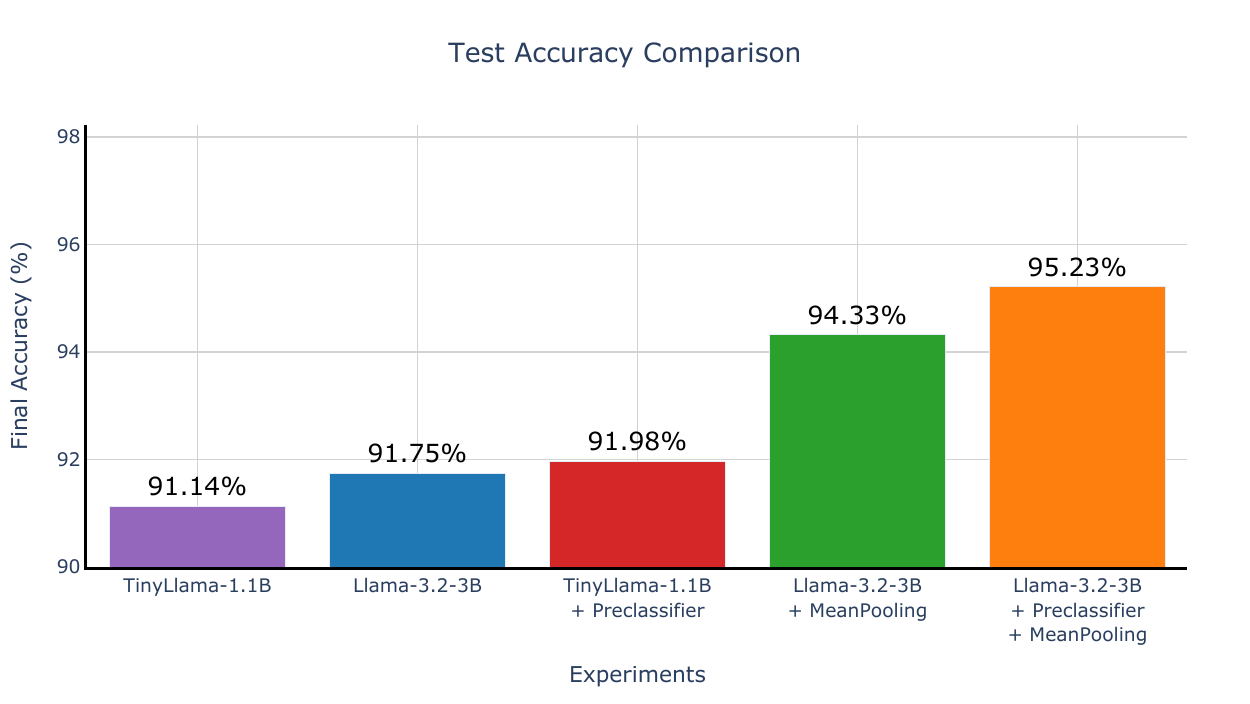}
    \caption{Using Llama-3.2-3B on AG news dataset with preclassifer and mean pooling.}
    \label{fig:exp:ag_news_cen}
\end{figure}

\begin{figure*}[t]
    \centering
    \begin{subfigure}[t]{0.49\textwidth}
        \centering
        \includegraphics[width=\linewidth]{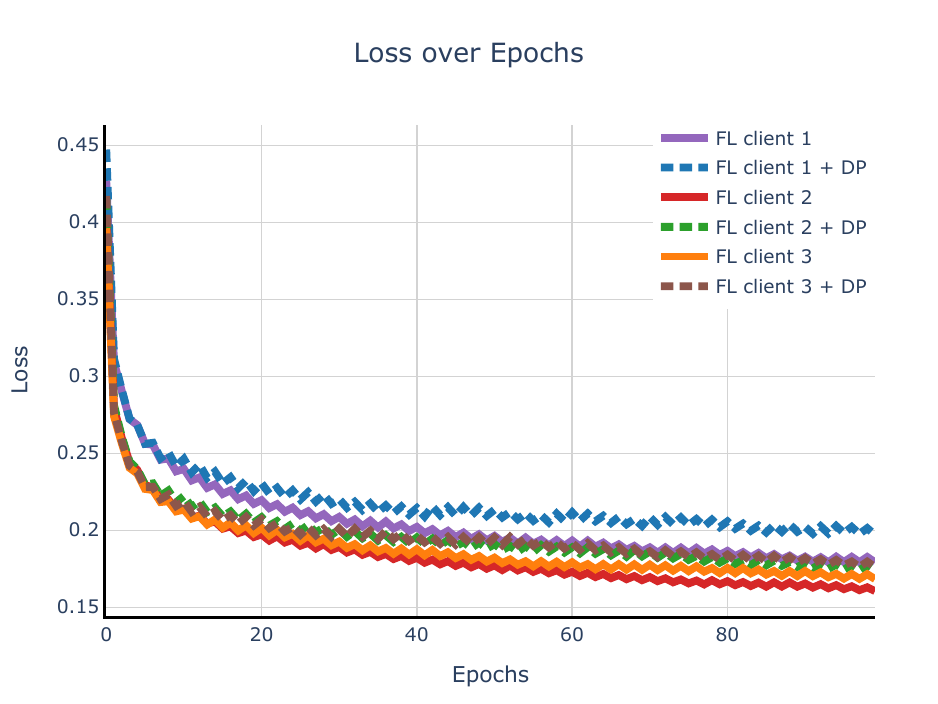}
        \caption{Loss for training the classifier head with mean pooling.}
    \end{subfigure}
    \hfill
    \begin{subfigure}[t]{0.49\textwidth}
        \centering
        \includegraphics[width=\linewidth]{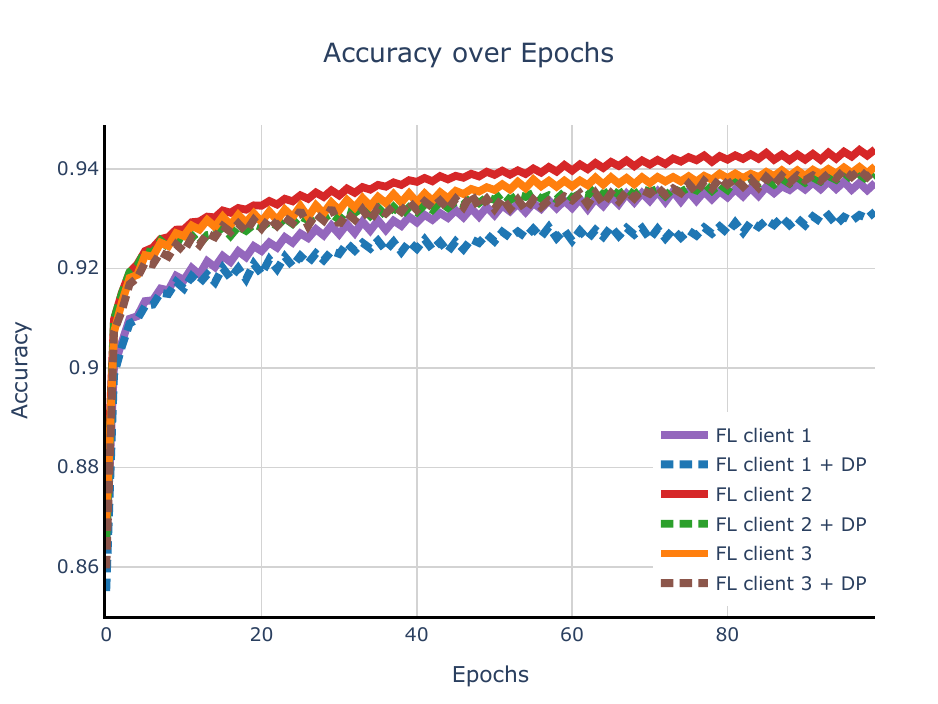}
        \caption{Accuracy for training the classifier head with mean pooling.}
    \end{subfigure}
    \caption{Federated training results for the AG news dataset with the Llama-3.2-3B model.}
    \label{fig:exp:ag_news_fl}
\end{figure*}

\begin{table}[t]
    {\small
    \setlength\tabcolsep{1.5pt}
    \centering
    \begin{tabular}{c|c|c|c|c}
        \toprule
        \multirow{3}{*}{Clients} & \multirow{3}{*}{Scenario} & Top-1 & Total train & \multirow{3}{*}{Speedup} \\
        & & accuracy & \& val time & \\
        & & (\%) & (min) & \\
        \midrule
        1 & Centralized & 94.18 & 10.81 & - \\
        \cline{1-5}
        \multirow{2}{*}{3} & FL & 93.39 & 4.55 & \textbf{2.38x}  \\
        \cline{2-5}
        & \begin{tabular}{c} FL + DP\\(adaptive) \end{tabular} & 92.92 & 4.57 & 2.37x \\
        \bottomrule
    \end{tabular}
    \caption{Summary of the results using Llama-3.2-3B on AG news dataset with mean pooling and adaptive DP with warmup.}
    \label{tab:ag_news}
    }
\end{table}

We first present our results on the AG news dataset. In Fig.~\ref{fig:exp:ag_news_cen}, we show that when training a classifier head on top of a frozen LLM, the best strategy is to use a larger LLM as the base model to obtain hidden states for classification, i.e., Llama-3.2-3B vs. TinyLlama-1.1B. Larger models offer greater complexity and produce richer embeddings and hidden state representations. If running the larger model locally is not feasible, this step can be delegated to a remote cluster as the LLM is frozen and only a single forward pass per prompt is needed to extract the hidden states. Fig.~\ref{fig:exp:ag_news_fl} further illustrates that to fully leverage the advantages of a larger model, computing the hidden states using the mean-pooling strategy, which aggregates information from all output tokens, is superior to just using the output EOS token. This approach better captures the model's representational power and improves classification performance. Additionally, smaller models such as TinyLlama-1.1B, can benefit from a pre-classifier layer which adds flexibility to the final classification stage and can lead to modest accuracy improvements.

Next, in Fig.~\ref{fig:exp:ag_news_fl} and Table~\ref{tab:ag_news}, we focus on training the best performing model for FL. Model convergence, accuracy, and training speedup are evaluated in a distributed setup using $4$ Mac Mini M4 devices, one serving as the aggregator and the rest as clients. To maximize training speed, the primary focus is on using three clients and evaluating the model’s accuracy both with and without local adaptive DP. This involves applying adaptive clipping and noise to the local models using a noise scale of $0.1$, clipping EMA decay of $0.9$, target clipping quantile of $0.9$, and a clipping update rate of $0.05$. The model under evaluation is Llama-3.2-3B-Instruct with a classifier head trained on hidden states obtained via mean-pooling. Each local client is trained for 25 rounds and 2 local epochs at each round. To measure speedup, distributed training performance is compared against the centralized case trained for 50 epochs.

As we see in Fig.~\ref{fig:exp:ag_news_fl} and Table~\ref{tab:ag_news}, the model completes training in 17 minutes and 58 seconds for the centralized training setup, reaching an accuracy of $94.39\%$. In comparison, although FL introduces a slight reduction of less than $1\%$ in accuracy, the three-client FL setup achieves greater training speed, completing the training in 7 minutes and 27 seconds, resulting in a $2.4$x speedup over the centralized baseline. Introducing adaptive DP in the FL setting adds privacy preserving noise and applies gradient clipping, both of which can restrict model updates and slow down convergence. We employ a warmup strategy to mitigate the early stage noise amplification introduced by adaptive DP, where 6 warmup epochs (equivalent to 3 rounds per client) are used before enabling adaptive DP. Our observations show that with warmup, the model converges more effectively, and it requires approximately twice the number of epochs to match the accuracy of a model trained without DP. 

\subsubsection{ArXiv metadata}

\begin{figure}[t]
    \centering
    \includegraphics[width=0.9\linewidth]{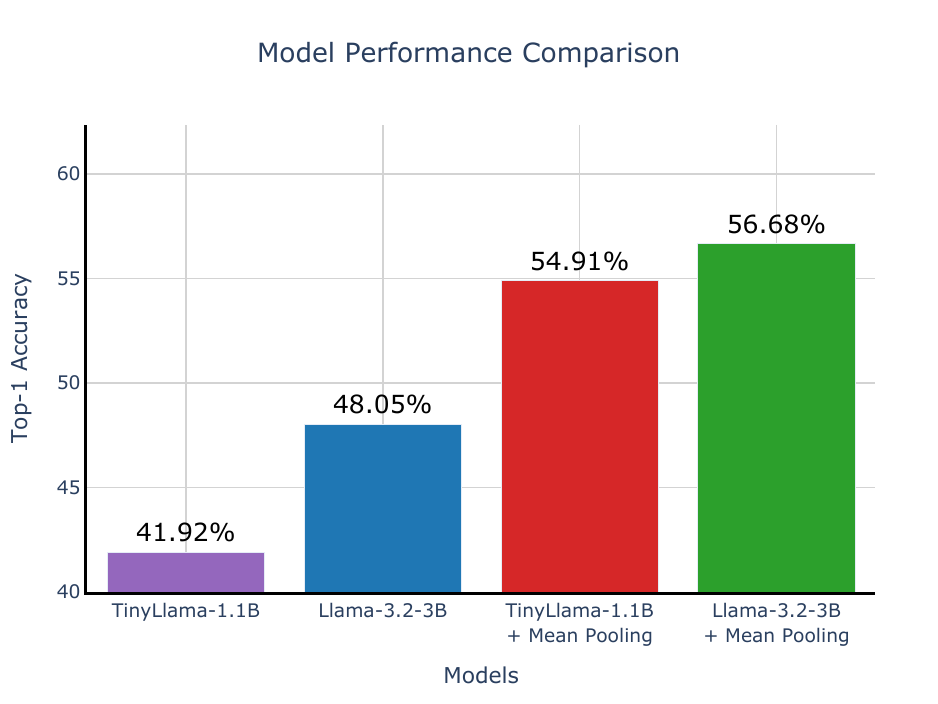}
    \caption{Using Llama-3.2-3B on arXiv metadata dataset.}
    \label{fig:exp:arxiv}
\end{figure}

Finally, we investigate our methodology on arXiv metadata, which is a larger dataset with $173$ classes. Similar to our findings in for AG news, Fig.~\ref{fig:exp:arxiv} illustrates that using a larger Llama-3.2-3B model compared to TinyLlama-1.1B results in better embeddings to train a classifier. Also, our approach can be further enhanced by employing mean pooling, which boosts the performance of both TinyLlama-1.1B and Llama-3.2-3B. We emphasize that we have reported top-1 accuracy for our methodology, and retrieving top-$K$ documents during retrieval with $K$ having a value around $3-10$ will result in a more accurate retrieval performance.

\section{Conclusion}

We proposed a novel retrieval architecture in our work, where instead of fully fine-tuning an encoder, we insert small adapter networks before the transformer blocks of an SLM and attach a classifier head to the model. The adapters and the classifier help the retriever to learn updated soft embeddings of a new corpus and train an accurate similarity mapping, respectively. Our paper enables two training schemes, where (i) training a classifier-head only requires significantly lower training time, while (ii) training both the adapter and the classifier results in a more accurate retrieval system. To bring down the training time, we further adopt FL and DP, which gives us a significant speedup proportional to the number of computing machines used.

\section{Acknowledgments}

The authors would like to express their sincere gratitude to the engineering team at webAI, especially Dr. Behnam Kia (Senior Director of AI) and Joel Dennison (Director of Innovation – R\&D), for their exceptional support, visionary leadership, and insightful discussions, all of which were instrumental in making this work possible.

\bibliography{aaai2026}

\onecolumn
\appendix

\section{Algorithm}

\begin{algorithm}[!ht]
    \caption{Training Classifier-as-Retriever with Soft-Embeddings using Federated Learning and Differential Privacy}
    \label{alg:alg}
    \DontPrintSemicolon
    \KwIn{$T, E, \{ \alpha_i, C_i^{(0)}, (\sigma_{0,i}^2)^{(0)}, \beta_i, \gamma_i, z_i \}_{i \in \mathcal{M}}$}
    \KwOut{$\theta_a^{(T+1)}, \theta_c^{(T+1)}$}

    \SetKwProg{Fn}{Function}{:}{}
    \SetKwFunction{FServer}{ServerUpdate}
    \Fn{\FServer{$T, E, K$}}{
        $t = 0$
        
        $\phi = \{ \phi_e, \phi_t, \phi_g \}$ \tcp*{Server loads pre-trained encoder $\phi_e$, transformers $\phi_t$ and generator $\phi_g$}
        
        $\theta^{(0)} = \{ \theta_a^{(0)}, \theta_c^{(0)} \}$ \tcp*{Server initializes adapter $\theta_a^{(0)}$ and classifier $\theta_c^{(0)}$}

        $\sigma_b = \frac{m}{20}$
        
        \ForAll{$i \in \mathcal{M}$}{
            $\phi_i = \phi$ \tcp*{Server broadcasts $\phi$ to clients}
            
            $\texttt{ClientInit}(i, T, \phi)$ \tcp*{Initialize fixed/adaptive DP parameters of clients}
        }
        
        \While{$t \le T$}{
            \ForAll{$i \in \mathcal{M}$}{
                $\theta_i^{(t)} = \theta^{(t)}$ \tcp*{Server broadcasts current global model $\theta^{(t)}$ to clients}
            
                $\theta_i^{(t+1)} = \texttt{ClientUpdate}(i, \theta^{(t)})$ \tcp*{Client $i$ runs local updates for $E$ epochs}
            }

            $\theta^{(t+1)} = \frac1m \sum_{i=1}^m{\theta_i^{(t+1)}}$ \tcp*{Aggregator averages all local models according to Eq.~\eqref{eqn:globalupdate}}
    
            $t \gets t+1$ \tcp*{Move to the next iteration}
        }
        
        \KwRet $\theta_a^{(T+1)}$, $\theta_c^{(T+1)}$
    }

    \SetKwFunction{FInit}{ClientInit}
    \Fn{\FInit{$i, \sigma_b$}}{    
        $z_{\Delta,i} = ( z_i^{-2} - (2 \sigma_b)^{-2} )^{-0.5}$

        \ForAll{$0 \le t \le T$}{
            \If{$\beta_i = 0$}{
                $C_i^{(t)} = C_i^{(0)}, \qquad ( \sigma_{0,i}^2 )^{(t)} = (\sigma_{0,i}^2)^{(0)} (C_i^{(0)})^2, \qquad ( \sigma_{1,i}^2 )^{(t)} = 0$ \tcp*{Fixed DP}
            } \Else{
                $( \sigma_{1,i}^2 )^{(t)} = 2 z_{\Delta,i}^2 \beta_i^2 \gamma_i^2$ \tcp*{Adaptive DP}
            }
        }
    }

    \SetKwFunction{FClient}{ClientUpdate}
    \Fn{\FClient{$i$, $\theta^{(t)}$}}{
        $\xi_i^{(t)} = (x_i^{(t)}, y_i^{(t)}) \sim \mathcal{D}_i$ \tcp*{Sample mini-batch}
        
        $\hat{y}^{(t)} = \Retriever_{\phi_e, \theta_a^{(t)}, \phi_t, \theta_c^{(t)}}(x_i^{(t)}, K)$ \tcp*{Forward pass $\xi_i^{(t)}$ according to Eq.~\eqref{eqn:retriever}}

        \tcp{Calculate local loss using Eq.~\eqref{eqn:objectives}, then run backpropagation}
        {\small $g_i(\theta^{(t)}, \xi_i^{(t)}) = \frac1{|\xi_i^{(t)}|} \sum_{j=1}^{|\xi_i^{(t)}|}{\nabla{\ell}(\Classifier_{\theta_c^{(t)}}(\Transformers_{\phi_t}(\Adapter_{\theta_a^{(t)}}(\Tokenize_{\phi_e}(x_j)))), y_j^{(t)})}$}

        $\theta_i^{(t+0.5)} = \theta^{(t)} - \alpha_i^{(t)} g_i(\theta^{(t)}, \xi_i^{(t)})$ \tcp*{Take stochastic gradient descent step}

        $\Delta \theta_i^{(t)} = \theta_i^{(t+0.5)} - \theta_i^{(t)}$ \tcp*{Calculate parameter update difference according to Eq.~\eqref{eqn:delta}}
        
        \tcp{Clip the gradient step and then add Gaussian noise according to Eq.~\eqref{eqn:localupdate}}
        $\theta_i^{(t+1)} = \theta_i^{(t)} + \min{( 1, \frac{C_i^{(t)}}{\| \Delta \theta_i^{(t)}} \| )} \Delta \theta_i^{(t)} + \mathcal{N}( \mathbf{0}, ( \sigma_{0,i}^2 )^{(t)} I + ( \sigma_{1,i}^2 )^{(t)} \|\Delta \theta_i^{(t)}\|^2 I )$

        \tcp{Update clipping thresholds and additive noise variances when adaptive DP is activated}
        \If{$\beta_i \neq 0$}{
            $C_i^{(t+1)} = (1 - \beta_i) C_i^{(t)} + \beta_i \gamma_i \| \Delta \theta_i^{(t)} \|$
                
            $( \sigma_{0,i}^2 )^{(t+1)} = 2 z_{\Delta,i}^2 (1 - \beta_i)^2 ( C_i^{(t+1)} )^2$
        }
        
        \KwRet $\theta_i^{(t+1)}$
    }
\end{algorithm}

\section{Lemmas and Proposition} \label{appendix:analysis}

In this section, we will build on the assumptions we made in Sec.~\ref{ssec:assump} to prove some Lemmas necessary to derive our main result in Theorem~\ref{theorem:fixed}.

\begin{lemma}[Global smoothness] \label{lemma:globalsmooth}
    Let Assumption~\ref{assump:smooth} hold. Then the global loss function $F(\theta)$ is $\bar{L}$-smooth, i.e., its gradient $\nabla{F}(\theta)$ is $\bar{L}$-Lipschitz continuous. We have
    \begin{equation}
        \norm{\nabla{F}(\theta) - \nabla{F}(\theta')} \le \bar{L} \norm{\theta - \theta'},
    \end{equation}
    where $\bar{L} = \frac1m \sum_{i=1}^m{L_i}$.

    \begin{proof}
        The proof follows from triangle inequality and then invoking Assumption~\ref{assump:smooth}. We have
        \begin{equation}
            \begin{aligned}
                \norm{\nabla{F}(\theta) - \nabla{F}(\theta')} & = \norm{\frac1m \sum_{i=1}^m{\left( \nabla{F}_i(\theta) - \nabla{F}_i(\theta') \right)}} \le \frac1m \sum_{i=1}^m{\norm{\nabla{F}_i(\theta) - \nabla{F}_i(\theta')}}
                \\
                & \le \frac1m \sum_{i=1}^m{L_i \norm{\theta - \theta'}} = \bar{L} \norm{\theta - \theta'}.
            \end{aligned}
        \end{equation}
    \end{proof}
\end{lemma}

Lemma~\ref{lemma:globalsmooth} proves that if the individual loss functions are smooth, then the global loss function which is a convex combination of the local losses, will also be smooth. We next present a key consequence of smooth functions.

\begin{lemma} \label{lemma:smoothupdate}
    Let Assumption~\ref{assump:smooth} hold. Then, Lemma~\ref{lemma:globalsmooth} implies that
    \begin{equation}
        F(\theta) \le F(\theta') + \nabla{F}(\theta')^T (\theta - \theta') + \frac12 \bar{L} \norm{\theta - \theta'}^2.
    \end{equation}

    \begin{proof}
        See \cite{bottou2018optimization} for the proof.
    \end{proof}
\end{lemma}

We next need to obtain an upper bound for the stochastic gradients of clients. We note that although in Assumption~\ref{assump:bound}, we assumed that the stochastic gradients are bounded, but in the next Lemma we will show that a less strict non-uniform bound can be derived for it as well.

\begin{lemma}[Stochastic gradient bound] \label{lemma:globalvar}
    Let Assumptions~\ref{assump:var} and \ref{assump:graddiv} hold. Then the average expected bound for stochastic gradients $g_i(\theta, \xi_i)$ is given as
    \begin{equation}
        \frac1m \sum_{i=1}^m{\bbEi{ \norm{g_i(\theta, \xi_i)}^2 }} \le 3 (\bar{\rho}_0^2 + \bar{\zeta}_0^2) + 3 (\bar{\rho}_1^2 + \bar{\zeta}_1^2 + 1) \norm{\nabla{F}(\theta)}^2.
    \end{equation}

    \begin{proof}
        We have
        \begin{equation}
            \begin{aligned}
                & \frac1m \sum_{i=1}^m{\bbEi{ \norm{g_i(\theta, \xi_i)}^2 }} = \frac1m \sum_{i=1}^m{\bbEi{ \norm{g_i(\theta, \xi_i) - \nabla{F}_i(\theta) + \nabla{F}_i(\theta) - \nabla{F}(\theta) + \nabla{F}(\theta)}^2 }}
                \\
                & \le \frac3m \sum_{i=1}^m{\bbEi{ \norm{g_i(\theta, \xi_i) - \nabla{F}_i(\theta)}^2}} + \frac3m \sum_{i=1}^m{\norm{\nabla{F}_i(\theta) - \nabla{F}(\theta)}^2} + \frac3m \sum_{i=1}^m{\norm{\nabla{F}(\theta)}^2}
                \\
                & \le 3 \left( \bar{\rho}_0^2 + \bar{\rho}_1^2 \norm{\nabla{F}(\theta)}^2 + \bar{\zeta}_0^2 + \bar{\zeta}_1^2 \norm{\nabla{F}(\theta)}^2 + \norm{\nabla{F}(\theta)}^2 \right)
                \\
                & = 3 (\bar{\rho}_0^2 + \bar{\zeta}_0^2) + 3 (\bar{\rho}_1^2 + \bar{\zeta}_1^2 + 1) \norm{\nabla{F}(\theta)}^2,
            \end{aligned}
        \end{equation}
        where we used Young's inequality first, and then invoked Assumptions~\ref{assump:var} and \ref{assump:graddiv}.
    \end{proof}
\end{lemma}

We will mainly use Lemma~\ref{lemma:globalvar} to bound the stochastic gradients in our subsequent analysis. The uniform bound of Assumption~\ref{assump:bound} will only be used to handle clipping of the gradients, as seen in the next Lemma.

\begin{lemma}[Loss decrement] \label{lemma:lossdec}
    Let Assumptions~\ref{assump:smooth}-\ref{assump:bound} hold. Then, Lemmas~\ref{lemma:smoothupdate} and \ref{lemma:globalvar} imply that
    \begin{equation}
        \begin{aligned}
            & \bbEt{F(\theta^{(t+1)})} \le F(\theta^{(t)})
            \\
            & - \left[ \mini{\alpha_i^{(t)}, \frac{C_i^{(t)}}{B_i}} - \frac32 \bar{L} \maxi{{\left( \alpha_i^{(t)} \right)}^2 \left( 1 + (\sigma_{1,i}^2)^{(t)} \right)} (\bar{\rho}_1^2 + \bar{\zeta}_1^2 + 1) \right] \norm{\nabla{F}(\theta^{(t)})}^2
            \\
            & + \frac12 \bar{L} \left( \overline{{(C^{(t)})}^2} + ( \overline{\sigma_0^2} )^{(t)} \right) + \frac32 \bar{L} \maxi{{\left( \alpha_i^{(t)} \right)}^2 \left( 1 + (\sigma_{1,i}^2)^{(t)} \right)} (\bar{\rho}_0^2 + \bar{\zeta}_0^2).
        \end{aligned}
    \end{equation}
    
    \begin{proof}
        Letting $\theta = \theta^{(t+1)}$ and $\theta' = \theta^{(t)}$ in Lemma~\ref{lemma:globalsmooth}, and taking the expected value of the whole inequality, we get
        \begin{equation}
            \begin{aligned}
                \bbEt{F(\theta^{(t+1)})} \le F(\theta^{(t)}) & + \nabla{F}(\theta^{(t)})^T \underbrace{\bbEt{\theta^{(t+1)} - \theta^{(t)}}}_{\triangleq T_1}
                \\
                & + \frac12 \bar{L} \underbrace{\bbEt{\norm{\theta^{(t+1)} - \theta^{(t)}}^2}}_{\triangleq T_2}.
            \end{aligned}
        \end{equation}
        First, focusing on the term $\theta^{(t+1)} - \theta^{(t)}$, we have that
        \begin{equation} \label{eqn:thetadiff}
            \theta^{(t+1)} - \theta^{(t)} = - \frac1m \sum_{i=1}^m{\min{\left( 1, \frac{C_i^{(t)}}{\norm{\Delta \theta_i^{(t)}}} \right)} \alpha_i^{(t)} g_i(\theta^{(t)}, \xi_i^{(t)})} + \frac1m \sum_{i=1}^m{\epsilon_i^{(t)}}.
        \end{equation}
        Now, noting that $\| \Delta \theta_i^{(t)} \| = \alpha_i^{(t)} \| g_i(\theta^{(t)}, \xi_i^{(t)}) \| \le \alpha_i^{(t)} B_i$, we next take the expected value of the above expression to obtain $T_1$ as
        \begin{equation} \label{eqn:T1}
            \begin{aligned}
                & \begin{aligned}
                    T_1 = \bbEt{\theta^{(t+1)} - \theta^{(t)}} = - \frac1m & \sum_{i=1}^m{\alpha_i^{(t)} \bbEit{ \min{\left( 1, \frac{C_i^{(t)}}{\norm{\Delta \theta_i^{(t)}}} \right)} g_i(\theta^{(t)}, \xi_i^{(t)})}}
                    \\
                    & + \frac1m \sum_{i=1}^m{\bbEt{\epsilon_i^{(t)}}}
                \end{aligned}
                \\
                & = - \frac1m \sum_{i=1}^m{\alpha_i^{(t)} \bbEit{ \min{\left( 1, \frac{C_i^{(t)}}{\norm{\Delta \theta_i^{(t)}}} \right)} g_i(\theta^{(t)}, \xi_i^{(t)})}} + \frac1m \sum_{i=1}^m{\bbEit{0}}
                \\
                & \le - \frac1m \sum_{i=1}^m{\alpha_i^{(t)} \min{\left( 1, \frac{C_i^{(t)}}{\alpha_i^{(t)} B_i} \right)} \bbEit{ g_i(\theta^{(t)}, \xi_i^{(t)})}} = - \frac1m \sum_{i=1}^m{\min{\left( \alpha_i^{(t)}, \frac{C_i^{(t)}}{B_i} \right)} \nabla{F}_i(\theta^{(t)})}
                \\
                & \le - \frac1m \mini{\alpha_i^{(t)}, \frac{C_i^{(t)}}{B_i}} \sum_{i=1}^m{\nabla{F}_i(\theta^{(t)})} = - \mini{\alpha_i^{(t)}, \frac{C_i^{(t)}}{B_i}} \nabla{F}(\theta^{(t)}).
            \end{aligned}
        \end{equation}
        As for the term $T_2$, we will first calculate $\| \theta^{(t)} - \theta^{(t)} \|^2$ using Eq.~\eqref{eqn:thetadiff} as
        \begin{equation} \label{eqn:thetanorm}
            \begin{aligned}
                \norm{\theta^{(t+1)} - \theta^{(t)}}^2 & = \norm{- \frac1m \sum_{i=1}^m{\min{\left( 1, \frac{C_i^{(t)}}{\norm{\Delta \theta_i^{(t)}}} \right)} \alpha_i^{(t)} g_i(\theta^{(t)}, \xi_i^{(t)})} + \frac1m \sum_{i=1}^m{\epsilon_i^{(t)}}}^2
                \\
                & \,\, \begin{aligned}
                    \le \, & \norm{- \frac1m \sum_{i=1}^m{\min{\left( 1, \frac{C_i^{(t)}}{\norm{\Delta \theta_i^{(t)}}} \right)} \alpha_i^{(t)} g_i(\theta^{(t)}, \xi_i^{(t)})}}^2 + \norm{\frac1m \sum_{i=1}^m{\epsilon_i^{(t)}}}^2
                    \\
                    & + 2 \dotproduct{- \frac1m \sum_{i=1}^m{\min{\left( 1, \frac{C_i^{(t)}}{\norm{\Delta \theta_i^{(t)}}} \right)} \alpha_i^{(t)} g_i(\theta^{(t)}, \xi_i^{(t)})}}{\frac1m \sum_{i=1}^m{\epsilon_i^{(t)}}}
                \end{aligned}
                \\
                & \,\, \begin{aligned}
                    \le \,  & \frac1m \sum_{i=1}^m{\min{\left( 1, \frac{C_i^{(t)}}{\norm{\Delta \theta_i^{(t)}}} \right)}^2 {\left( \alpha_i^{(t)} \right)}^2 \norm{g_i(\theta^{(t)}, \xi_i^{(t)})}^2} + \frac1m \sum_{i=1}^m{\norm{\epsilon_i^{(t)}}^2}
                    \\
                    & - \frac2{m^2} \sum_{i=1}^m{\sum_{j=1}^m{\min{\left( 1, \frac{C_i^{(t)}}{\norm{\Delta \theta_i^{(t)}}} \right)} \alpha_i^{(t)} \dotproduct{g_i(\theta^{(t)}, \xi_i^{(t)})}{\epsilon_j^{(t)}}}}.
                \end{aligned}
            \end{aligned}
        \end{equation}
        Noting that  $\min( 1, \frac{C_i^{(t)}}{\| \Delta \theta_i^{(t)} \|} )^2 ( \alpha_i^{(t)} )^2 \| g_i(\theta^{(t)}, \xi_i^{(t)}) \|^2 = \min( \| \Delta \theta_i^{(t)} \|, C_i^{(t)} )^2 \le \| \Delta \theta_i^{(t)} \|^2 + (C_i^{(t)})^2$, and that $\mathbb{E}_{\epsilon_j^{(t)} \sim \mathcal{N}_j^{(t)}}{[ \epsilon_j^{(t)} ]} = 0$, we take the expected value of Eq.~\eqref{eqn:thetanorm} to obtain $T_2$ as
        \begin{equation}
            \begin{aligned}
                T_2 & = \bbEt{\norm{\theta^{(t+1)} - \theta^{(t)}}^2}
                \\
                & \le \frac1m \sum_{i=1}^m{{\left( \alpha_i^{(t)} \right)}^2 \bbEit{\norm{g_i(\theta^{(t)}, \xi_i^{(t)})}^2}} + \frac1m \sum_{i=1}^m{{(C_i^{(t)})}^2} + \frac1m \sum_{i=1}^m{\bbEt{\norm{\epsilon_i^{(t)}}^2}}
                \\
                & \begin{aligned}
                    \,\, = \frac1m \sum_{i=1}^m{{\left( \alpha_i^{(t)} \right)}^2 \bbEit{\norm{g_i(\theta^{(t)}, \xi_i^{(t)})}^2}} & + \frac1m \sum_{i=1}^m{{(C_i^{(t)})}^2}
                    \\
                    & + \frac1m \sum_{i=1}^m{\bbEit{( \sigma_{0,i}^2 )^{(t)} + (\sigma_{1,i}^2)^{(t)} \norm{\Delta \theta_i^{(t)}}^2}}
                \end{aligned}
                \\
                & = \frac1m \sum_{i=1}^m{\left( {(C_i^{(t)})}^2 + ( \sigma_{0,i}^2 )^{(t)} \right)} + \frac1m \sum_{i=1}^m{{\left( \alpha_i^{(t)} \right)}^2 \left( 1 + (\sigma_{1,i}^2)^{(t)} \right) \bbEit{\norm{g_i(\theta^{(t)}, \xi_i^{(t)})}^2}}
                \\
                & \begin{aligned}
                    \,\, = \frac1m & \sum_{i=1}^m{\left( {(C_i^{(t)})}^2 + ( \sigma_{0,i}^2 )^{(t)} \right)}
                    \\
                    & + 3 \maxi{{\left( \alpha_i^{(t)} \right)}^2 \left( 1 + (\sigma_{1,i}^2)^{(t)} \right)} \left( (\bar{\rho}_0^2 + \bar{\zeta}_0^2) + (\bar{\rho}_1^2 + \bar{\zeta}_1^2 + 1) \norm{\nabla{F}(\theta^{(t)})}^2 \right).
                \end{aligned}
            \end{aligned}
        \end{equation}
        Combining $T_1$ and $T_2$ concludes the proof.
    \end{proof}
\end{lemma}

In Lemma~\ref{lemma:lossdec}, we showed how much the global loss is changed in one iteration of training. We observe that in one iteration of training, the global loss $F(\theta^{(t)})$ is decremented by a scaled version of $\| \nabla{F}(\theta^{(t)}) \|^2$, and incremented by two noise terms, one depending on clipping and another on the gradient variances. In the next proposition, we obtain condition under which the resulting upper bound for $F(\theta^{(t+1)})$ gets reduced at every iteration of training, i.e., the decrement on the loss mentioned above dominates the increment based on additive noises.

\begin{proposition}[Final loss] \label{prop:finalloss}
    Let Assumptions~\ref{assump:smooth}-\ref{assump:bound} hold. Also, for all clients $i \in \mathcal{M}$ and all iterations $0 \le t \le T$, let the learning rate $\alpha_i^{(t)}$ satisfy
    \begin{equation}
        \begin{gathered}
            \alpha_i^{(t)} < \sqrt{\frac2{3 \bar{L} (\bar{\rho}_1^2 + \bar{\zeta}_1^2 + 1) \Gamma_1} \frac{\mini{\frac{C_i^{(t)}}{B_i}}}{\maxi{1 + (\sigma_{1,i}^2)^{(t)}}}},
            \\
            \frac{\maxi{\alpha_i^{(t)}}^2}{\mini{\alpha_i^{(t)}}} < \frac2{3 \bar{L} (\bar{\rho}_1^2 + \bar{\zeta}_1^2 + 1) \Gamma_2} \frac1{\maxi{1 + (\sigma_{1,i}^2)^{(t)}}},
        \end{gathered}
    \end{equation}
    in which $\Gamma_1 > 1$ and $\Gamma_2 > 1$ are two arbitrary real numbers. Then, Lemma~\ref{lemma:lossdec} implies that
    \begin{equation}
        \begin{aligned}
            \bbE{F(\theta^{(T+1)})} \le F(\theta^{(0)}) & - \sum_{t=0}^T{\omega^{(t)} \norm{\nabla{F}(\theta^{(t)})}^2} + \frac12 \bar{L} \sum_{t=0}^T{\left( \overline{{(C^{(t)})}^2} + ( \overline{\sigma_0^2} )^{(t)} \right)}
            \\
            & + \frac32 \bar{L} (\bar{\rho}_0^2 + \bar{\zeta}_0^2) \sum_{t=0}^T{\maxi{ {\left( \alpha_i^{(t)} \right)}^2 \left( 1 + (\sigma_{1,i}^2)^{(t)} \right)}}.
        \end{aligned}
    \end{equation}
    where $\omega^{(t)} = \min_{i \in \mathcal{M}}{( \frac{C_i^{(t)}}{B_i} ( 1 - \frac1{\Gamma_1} ), \alpha_i^{(t)} ( 1 - \frac1{\Gamma_2} ) )}$.
    
    \begin{proof}
        Recursively expanding Lemma~\ref{lemma:lossdec} gives us
        \begin{equation} \label{eqn:lossdec_explicit}
            \begin{aligned}
                & \bbE{F(\theta^{(T+1)})} \le F(\theta^{(0)})
                \\
                & - \sum_{t=0}^T{\left[ \mini{\alpha_i^{(t)}, \frac{C_i^{(t)}}{B_i}} - \frac32 \bar{L} \maxi{{\left( \alpha_i^{(t)} \right)}^2 \left( 1 + (\sigma_{1,i}^2)^{(t)} \right)} (\bar{\rho}_1^2 + \bar{\zeta}_1^2 + 1) \right] \norm{\nabla{F}(\theta^{(t)})}^2}
                \\
                & + \frac12 \bar{L} \sum_{t=0}^T{\left( \overline{{(C^{(t)})}^2} + ( \overline{\sigma_0^2} )^{(t)} \right)} + \frac32 \bar{L} (\bar{\rho}_0^2 + \bar{\zeta}_0^2) \sum_{t=0}^T{\maxi{{\left( \alpha_i^{(t)} \right)}^2 \left( 1 + (\sigma_{1,i}^2)^{(t)} \right)}}.
            \end{aligned}
        \end{equation}
        To ensure that the loss is decremented at every iteration, a necessary condition is to have
        \begin{equation} \label{eqn:omega}
            \omega^{(t)} = \mini{\alpha_i^{(t)}, \frac{C_i^{(t)}}{B_i}} - \frac32 \bar{L} \maxi{{\left( \alpha_i^{(t)} \right)}^2 \left( 1 + (\sigma_{1,i}^2)^{(t)} \right)} (\bar{\rho}_1^2 + \bar{\zeta}_1^2 + 1) > 0,
        \end{equation}
        which holds if
        \begin{equation}
            \frac{\mini{\alpha_i^{(t)}, \frac{C_i^{(t)}}{B_i}}}{\maxi{{\left( \alpha_i^{(t)} \right)}^2 \left( 1 + (\sigma_{1,i}^2)^{(t)} \right)}} > \frac32 \bar{L} (\bar{\rho}_1^2 + \bar{\zeta}_1^2 + 1).
        \end{equation}
        To simplify this condition, let us break the maximum function in the denominator as
        \begin{equation}
            \frac{\mini{\alpha_i^{(t)}, \frac{C_i^{(t)}}{B_i}}}{\maxi{\alpha_i^{(t)}}^2} > \frac32 \bar{L} (\bar{\rho}_1^2 + \bar{\zeta}_1^2 + 1) \maxi{1 + (\sigma_{1,i}^2)^{(t)}}.
        \end{equation}
        Let us analyze the possibilities case by case.
        \begin{itemize}
            \item $\exists j \in \mathcal{M}$ s.t. $\frac{C_j^{(t)}}{B_j} < \min_{i \in \mathcal{M}}{( \alpha_i^{(t)} )}$: This corresponds to the case where clipping occurs, which gives us
            \begin{equation}
                \maxi{\alpha_i^{(t)}} < \sqrt{\frac2{3 \bar{L} (\bar{\rho}_1^2 + \bar{\zeta}_1^2 + 1)} \frac{\mini{\frac{C_i^{(t)}}{B_i}}}{\maxi{1 + (\sigma_{1,i}^2)^{(t)}}}}.
            \end{equation}
            Let us define a variable $\Gamma_1$ which satisfies $\Gamma_1 > 1$, and rewrite the constraint above as
            \begin{equation}
                \maxi{\alpha_i^{(t)}} < \sqrt{\frac2{3 \bar{L} (\bar{\rho}_1^2 + \bar{\zeta}_1^2 + 1) \Gamma_1} \frac{\mini{\frac{C_i^{(t)}}{B_i}}}{\maxi{1 + (\sigma_{1,i}^2)^{(t)}}}}.
            \end{equation}
            This will result in the following lower bounds for $\omega^{(t)}$ defined in Eq.~\eqref{eqn:omega}
            \begin{equation}
                \begin{cases}
                    \omega^{(t)} > \frac32 \bar{L} (\Gamma_1 - 1) (\bar{\rho}_1^2 + \bar{\zeta}_1^2 + 1) \maxi{1 + (\sigma_{1,i}^2)^{(t)}}^2 \maxi{\alpha_i^{(t)}}^2,
                    \\
                    \omega^{(t)} > \mini{\frac{C_i^{(t)}}{B_i}} \left( 1 - \frac1{\Gamma_1} \right).
                \end{cases}
            \end{equation}
            
            \item $\exists j \in \mathcal{M}$ s.t. $\alpha_j^{(t)} < \min_{i \in \mathcal{M}}{( \frac{C_i^{(t)}}{B_i} )}$: This is the case where no clipping occurs, which results in
            \begin{equation}
                \frac{\maxi{\alpha_i^{(t)}}^2}{\mini{\alpha_i^{(t)}}} < \frac2{3 \bar{L} (\bar{\rho}_1^2 + \bar{\zeta}_1^2 + 1)} \frac1{\maxi{1 + (\sigma_{1,i}^2)^{(t)}}}.
            \end{equation}
            Let us define a variable $\Gamma_2$ which satisfies $\Gamma_2 > 1$. Using this definition, the above constraint can be rewritten as
            \begin{equation}
                \frac{\maxi{\alpha_i^{(t)}}^2}{\mini{\alpha_i^{(t)}}} < \frac2{3 \bar{L} (\bar{\rho}_1^2 + \bar{\zeta}_1^2 + 1) \Gamma_2} \frac1{\maxi{1 + (\sigma_{1,i}^2)^{(t)}}}.
            \end{equation}
            This will result in the following lower bounds for $\omega^{(t)}$ defined in Eq.~\eqref{eqn:omega}
            \begin{equation}
                \begin{cases}
                    \omega^{(t)} > \frac{3 \bar{L} (\bar{\rho}_1^2 + \bar{\zeta}_1^2 + 1)}2 \maxi{1 + (\sigma_{1,i}^2)^{(t)}} \left( \Gamma_2 - 1 \right) \maxi{\alpha_i^{(t)}}^2,
                    \\
                    \omega^{(t)} > \mini{\alpha_i^{(t)}} \left( 1 - \frac1{\Gamma_2} \right).
                \end{cases}
            \end{equation}
        \end{itemize}
    \end{proof}
\end{proposition}

\section{Adaptive DP Analysis} \label{appendix:adaptive}

In adaptive DP, the clipping threshold and the noise variance are adapted at every iteration. Let $\sigma_b$ denote the noise standard deviation on the number of clients doing gradient clipping at a given iteration, which is heuristically taken to be $\sigma_b = \frac{m}{20}$ with $m$ being the total number of clients. Letting $z_i$ be a fixed noise multiplier for client $i \in \mathcal{M}$, define $z_{\Delta,i} = ( z_i^{-2} - (2 \sigma_b)^{-2} )^{-0.5}$ as the noise multiplier on parameter difference updates, i.e., $\Delta \theta_i$. Having defined this multiplier, the standard deviation of the noise that is added to the model parameters will be calculated as $z_{\Delta,i} C_i$. To adaptively change $C_i$, we will clip the gradients to a value at a specified target quantile of the update norm distribution, denoted as $\gamma_i$. Note that adapting $C_i$ will automatically adapt the noise variance $z_{\Delta,i} C_i$, therefore, the only adaptive parameter in our DP approach will be $C_i$. Letting $\beta_i$ denote the rate at which $C_i$ is modified, we will have
\begin{equation}
    C_i = (1 - \beta_i) C_i + \beta_i \gamma_i \norm{\Delta \theta_i}.
\end{equation}

Let us first write the iterate relations for adaptive DP, similar to what we did in Sec.~\ref{ssec:updates}. The clipping threshold will be updated as
\begin{equation} \label{eqn:clipupdate}
    C_i^{(t+1)} = (1 - \beta_i) C_i^{(t)} + \beta_i \gamma_i \norm{\Delta \theta_i^{(t)}}.
\end{equation}
Subsequently, the noise variance will be computed as $z_{\Delta,i}^2 ( C_i^{(t)} )^2$. However, to get this variance in the format of Eq.~\eqref{eqn:noise}, we want to avoid the term $\| \Delta \theta_i^{(t)} \|$ when taking the square of $C_i^{(t)}$. We thus obtain an upper bound for $( \sigma_{0,i}^2 )^{(t)}$ and $( \sigma_{1,i}^2 )^{(t)}$ as
\begin{equation} \label{eqn:noiseupdate}
    ( \sigma_{0,i}^2 )^{(t)} \le 2 z_{\Delta,i}^2 (1 - \beta_i)^2 {\left( C_i^{(t)} \right)}^2, \qquad ( \sigma_{1,i}^2 )^{(t)} \le 2 z_{\Delta,i}^2 \beta_i^2 \gamma_i^2.
\end{equation}
We now make an assumption for the expected behavior of clipping in the clients.

\begin{assumption} \label{assump:clipnum}
    Let $\nu_i^{(t)} \in \{ 0, 1 \}$ denote the indicator function that indicates whether clipping occurs at client $i \in \mathcal{M}$ at iteration $t$. We assume that
    \begin{equation}
        \mathbb{E}_{\nu_i^{(t)} \sim \mathcal{B}(1, p_i)}{\left[ \nu_i^{(t)} \right]} = p_i,
    \end{equation}
    where $0 \le p_i < 1$. Also, define $p = \maxi{p_i}$ with $0 \le p < 1$.
\end{assumption}
Assumption~\ref{assump:clipnum} essentially ensures that there are infinitely many iterations (with probability $1 - p_i$ since $p_i < 1$) that a client perform regular gradient descent without clipping the gradients.

\begin{lemma}[Loss decrement] \label{lemma:lossdec_adaptive}
    Let Assumptions~\ref{assump:smooth}-\ref{assump:bound} and \ref{assump:clipnum} hold. Then, Lemmas~\ref{lemma:lossdec} implies that
    \begin{equation}
        \begin{aligned}
            & \mathbb{E}_{\substack{\{ \xi_i^{(t)} \sim \mathcal{D}_i \}_{i \in \mathcal{M}} \\ \{ \epsilon_i^{(t)} \sim \mathcal{N}_i^{(t)} \}_{i \in \mathcal{M}} \\ \{ \nu_i^{(t)} \sim \mathcal{B}(1, p_i) \}_{i \in \mathcal{M}}} }{\left[ F(\theta^{(t+1)}) \right]} \le F(\theta^{(t)})
            \\
            & - \left[ \mini{\alpha_i^{(t)}} {(1 - \maxi{p_i})}^m - \frac32 \bar{L} \maxi{{\left( \alpha_i^{(t)} \right)}^2 \left( 1 + (\sigma_{1,i}^2)^{(t)} \right)} (\bar{\rho}_1^2 + \bar{\zeta}_1^2 + 1) \right] \norm{\nabla{F}(\theta^{(t)})}^2
            \\
            & + \frac12 \bar{L} \left( \overline{{(C^{(t)})}^2} + ( \overline{\sigma_0^2} )^{(t)} \right) + \frac32 \bar{L} \maxi{{\left( \alpha_i^{(t)} \right)}^2 \left( 1 + (\sigma_{1,i}^2)^{(t)} \right)} (\bar{\rho}_0^2 + \bar{\zeta}_0^2).
        \end{aligned}
    \end{equation}
    
    \begin{proof}
        Starting from the result of Lemma~\ref{lemma:lossdec}, we have that
        \begin{equation}
            \begin{aligned}
                & \bbEt{F(\theta^{(t+1)})} \le F(\theta^{(t)})
                \\
                & - \left[ \mini{\alpha_i^{(t)}, \frac{C_i^{(t)}}{B_i}} - \frac32 \bar{L} \maxi{{\left( \alpha_i^{(t)} \right)}^2 \left( 1 + (\sigma_{1,i}^2)^{(t)} \right)} (\bar{\rho}_1^2 + \bar{\zeta}_1^2 + 1) \right] \norm{\nabla{F}(\theta^{(t)})}^2
                \\
                & + \frac12 \bar{L} \left( \overline{{(C^{(t)})}^2} + ( \overline{\sigma_0^2} )^{(t)} \right) + \frac32 \bar{L} \maxi{{\left( \alpha_i^{(t)} \right)}^2 \left( 1 + (\sigma_{1,i}^2)^{(t)} \right)} (\bar{\rho}_0^2 + \bar{\zeta}_0^2)
            \end{aligned}
        \end{equation}
        Focusing on the term $- \min_{i \in \mathcal{M}}{(\alpha_i^{(t)}, \frac{C_i^{(t)}}{B_i})}$, we can obtain an upper bound for it as
        \begin{equation}
            \begin{aligned}
                - \mini{\alpha_i^{(t)}, \frac{C_i^{(t)}}{B_i}} & = - \mini{\alpha_i^{(t)} \left( 1 - \nu_i^{(t)} \right) + \frac{C_i^{(t)}}{B_i} \nu_i^{(t)}}
                \\
                & \le - \left( \mini{\alpha_i^{(t)}} \left( 1 - \maxi{\nu_i^{(t)}} \right) + \mini{\frac{C_i^{(t)}}{B_i}} \mini{\nu_i^{(t)}} \right)
                \\
                & \le - \mini{\alpha_i^{(t)}} \left( 1 - \maxi{\nu_i^{(t)}} \right) = - \mini{\alpha_i^{(t)}} \mini{1 - \nu_i^{(t)}}
            \end{aligned}
        \end{equation}
        Now, taking its expected value of the above expression with respect to the random indicator variables $v_i^{(t)}$, we have that
        \begin{equation}
            \begin{aligned}
                \mathbb{E}_{\{ \nu_i^{(t)} \sim B(1, p_i) \}_{i \in \mathcal{M}}} & {\left[ - \mini{\alpha_i^{(t)}, \frac{C_i^{(t)}}{B_i}} \right]} \le - \mini{\alpha_i^{(t)}} \mathbb{E}_{\{ \nu_i^{(t)} \sim B(1, p_i) \}_{i \in \mathcal{M}}}{\left[ \mini{1 - \nu_i^{(t)}} \right]}
                \\
                & = - \mini{\alpha_i^{(t)}} \prod_{i=1}^m{(1 - p_i)} = - \mini{\alpha_i^{(t)}} {(1 - \maxi{p_i})}^m.
            \end{aligned}
        \end{equation}
        The proof is concluded by substituting the above expression back in Lemma~\ref{lemma:lossdec}.
    \end{proof}
\end{lemma}

\begin{proposition}[Final loss] \label{prop:finalloss_adaptive}
    Let Assumptions~\ref{assump:smooth}-\ref{assump:bound} and \ref{assump:clipnum} hold. Also, for all clients $i \in \mathcal{M}$ and all iterations $0 \le t \le T$, let the learning rate $\alpha_i^{(t)}$ satisfy
    \begin{equation}
        \frac{\maxi{\alpha_i^{(t)}}^2}{\mini{\alpha_i^{(t)}}} < \frac{2{(1 - \maxi{p_i})}^m}{3 \bar{L} (\bar{\rho}_1^2 + \bar{\zeta}_1^2 + 1) \Gamma_2} \frac1{\maxi{1 + (\sigma_{1,i}^2)^{(t)}}}
    \end{equation}
    in which $\Gamma > 1$ is an arbitrary real number. Then, Lemma~\ref{lemma:lossdec_adaptive} implies that
    \begin{equation}
        \begin{aligned}
            \bbE{F(\theta^{(T+1)})} \le F(\theta^{(0)}) & - \omega \sum_{t=0}^T{\mini{\alpha_i^{(t)}} \norm{\nabla{F}(\theta^{(t)})}^2} + \frac12 \bar{L} \sum_{t=0}^T{\left( \overline{{(C^{(t)})}^2} + ( \overline{\sigma_0^2} )^{(t)} \right)}
            \\
            & + \frac32 \bar{L} (\bar{\rho}_0^2 + \bar{\zeta}_0^2) \sum_{t=0}^T{\maxi{ {\left( \alpha_i^{(t)} \right)}^2 \left( 1 + (\sigma_{1,i}^2)^{(t)} \right)}}.
        \end{aligned}
    \end{equation}
    where $\omega = {(1 - \maxi{p_i})}^m \left( 1 - \frac1{\Gamma} \right)$.
    
    \begin{proof}
        Recursively expanding Lemma~\ref{lemma:lossdec_adaptive} gives us
        \begin{equation} \label{eqn:lossdec_explicit_adaptive}
            \begin{aligned}
                & \bbE{F(\theta^{(T+1)})} \le F(\theta^{(0)})
                \\
                & - \sum_{t=0}^T{\left[ \mini{\alpha_i^{(t)}} {(1 - \maxi{p_i})}^m - \frac32 \bar{L} \maxi{{\left( \alpha_i^{(t)} \right)}^2 \left( 1 + (\sigma_{1,i}^2)^{(t)} \right)} (\bar{\rho}_1^2 + \bar{\zeta}_1^2 + 1) \right] \norm{\nabla{F}(\theta^{(t)})}^2}
                \\
                & + \frac12 \bar{L} \sum_{t=0}^T{\left( \overline{{(C^{(t)})}^2} + ( \overline{\sigma_0^2} )^{(t)} \right)} + \frac32 \bar{L} (\bar{\rho}_0^2 + \bar{\zeta}_0^2) \sum_{t=0}^T{\maxi{{\left( \alpha_i^{(t)} \right)}^2 \left( 1 + (\sigma_{1,i}^2)^{(t)} \right)}}.
            \end{aligned}
        \end{equation}
        To ensure that the loss is decremented at every iteration, a necessary condition is to have
        \begin{equation} \label{eqn:omega_adaptive}
            \omega^{(t)} = \mini{\alpha_i^{(t)}} {(1 - \maxi{p_i})}^m - \frac32 \bar{L} \maxi{{\left( \alpha_i^{(t)} \right)}^2 \left( 1 + (\sigma_{1,i}^2)^{(t)} \right)} (\bar{\rho}_1^2 + \bar{\zeta}_1^2 + 1) > 0,
        \end{equation}
        which holds if
        \begin{equation}
            \frac{\mini{\alpha_i^{(t)}}}{\maxi{{\left( \alpha_i^{(t)} \right)}^2 \left( 1 + (\sigma_{1,i}^2)^{(t)} \right)}} > \frac{3 \bar{L} (\bar{\rho}_1^2 + \bar{\zeta}_1^2 + 1)}{2{(1 - \maxi{p_i})}^m}.
        \end{equation}
        To simplify this condition, let us break the maximum function in the denominator as
        \begin{equation}
            \frac{\mini{\alpha_i^{(t)}}}{\maxi{\alpha_i^{(t)}}^2} > \frac{3 \bar{L} (\bar{\rho}_1^2 + \bar{\zeta}_1^2 + 1)}{2{(1 - \maxi{p_i})}^m} \maxi{1 + (\sigma_{1,i}^2)^{(t)}}.
        \end{equation}
        Let us define a variable $\Gamma$ which satisfies $\Gamma > 1$, and rewrite the constraint above as
        \begin{equation}
            \frac{\maxi{\alpha_i^{(t)}}^2}{\mini{\alpha_i^{(t)}}} < \frac{2{(1 - \maxi{p_i})}^m}{3 \bar{L} (\bar{\rho}_1^2 + \bar{\zeta}_1^2 + 1) \Gamma_2} \frac1{\maxi{1 + (\sigma_{1,i}^2)^{(t)}}}.
        \end{equation}
        This will result in the following lower bound for $\omega^{(t)}$ defined in Eq.~\eqref{eqn:omega_adaptive}
        \begin{equation}
            \omega^{(t)} > {(1 - \maxi{p_i})}^m \mini{\alpha_i^{(t)}} \left( 1 - \frac1{\Gamma} \right).
        \end{equation}
        Factoring out $\min_{i \in \mathcal{M}}{(\alpha_i^{(t)})}$ from $\omega^{(t)}$ concludes the proof.
    \end{proof}
\end{proposition}

\begin{lemma} \label{lemma:recursive}
    Let Assumption~\ref{assump:bound} hold, and $C_i^{(t+1)} = (1 - \beta_i) C_i^{(t)} + \beta_i \gamma_i \| \Delta \theta_i^{(t)} \|$ be the formula for dynamically updating the clipping threshold with a constant learning rate $\alpha_i^{(t)} = \alpha$ for all clients $i \in \mathcal{M}$. We then have
    \begin{enumerate}[label=(\roman*)]
        \item $C_i^{(t)} \le {(1 - \beta_i)}^t C_i^{(0)} + \gamma_i B_i \alpha$, \label{lemma:recursive:C}

        \item ${\left( C_i^{(t)} \right)}^2 \le {(1 - \beta_i)}^{2t} {\left( C_i^{(0)} \right)}^2 + 2 {(1 - \beta_i)}^t \gamma_i B_i \alpha C_i^{(0)} + \gamma_i^2 B_i^2 \alpha^2$, \label{lemma:recursive:C2}

        \item $\sum_{t=0}^T{\overline{{(C^{(t)})}^2}} \le \frac{{\left( C^{(0)} \right)}^2}{1 - (1 - \mini{\beta_i})^2} + \frac{2 \maxi{\gamma_i B_i} \alpha \bar{C}^{(0)}}{\mini{\beta_i}} + \maxi{\gamma_i B_i}^2 \alpha^2 (T+1)$. \label{lemma:recursive:sumC2}
    \end{enumerate}

    \begin{proof}
        \ref{lemma:recursive:C} Using the update formula for the clipping threshold, we have
        \begin{equation}
            \begin{aligned}
                C_i^{(t)} & = (1 - \beta_i) C_i^{(t-1)} + \beta_i \gamma_i \norm{\Delta \theta_i^{(t-1)}} = {(1 - \beta_i)}^t C_i^{(0)} + \beta_i \gamma_i \sum_{u=0}^{t-1}{{(1-\beta_i)}^{t-1-u} \norm{\Delta \theta_i^{(u)}}}
                \\
                & \le {(1 - \beta_i)}^t C_i^{(0)} + \beta_i \gamma_i \alpha B_i \sum_{u=0}^\infty{{(1-\beta_i)}^{t-1-u}} \le {(1 - \beta_i)}^t C_i^{(0)} + \frac{\beta_i \gamma_i \alpha B_i}{1 - (1 - \beta_i)},
            \end{aligned}
        \end{equation}
        where we used $\| \Delta \theta_i^{(t)} \| = \alpha_i^{(t)} \| g_i(\theta^{(t)}, \xi_i^{(t)}) \| \le \alpha B_i$.

        \ref{lemma:recursive:C2} This can be easily proved by using the result of part~\ref{lemma:recursive:C} and squaring it.

        \ref{lemma:recursive:sumC2} Since we have $( C^{(t+1)} )^2 = \frac1m{\sum_{i=1}^m{( C_i^{(t+1)} )^2}}$ by definition, we have
        \begin{equation}
            \begin{aligned}
                \sum_{t=0}^T{\overline{{(C^{(t)})}^2}} = \sum_{t=0}^T{\frac1m{\sum_{i=1}^m{{\left( C_i^{(t+1)} \right)}^2}}} = \frac1m{\sum_{i=1}^m{\sum_{t=0}^T{{\left( C_i^{(t+1)} \right)}^2}}}
            \end{aligned}
        \end{equation}
        We can now use the results of part~\ref{lemma:recursive:C2} to write
        \begin{equation}
            \begin{aligned}
                \sum_{t=0}^T{{\left( C_i^{(t)} \right)}^2} & \le \sum_{t=0}^T{\left( {(1 - \beta_i)}^{2t} {\left( C_i^{(0)} \right)}^2 + 2 {(1 - \beta_i)}^t \gamma_i B_i \alpha C_i^{(0)} + \gamma_i^2 B_i^2 \alpha^2 \right)}
                \\
                & \le {\left( C_i^{(0)} \right)}^2 \sum_{t=0}^\infty{{(1 - \beta_i)}^{2t}} + 2 \gamma_i B_i \alpha C_i^{(0)} \sum_{t=0}^\infty{{(1 - \beta_i)}^t} + \gamma_i^2 B_i^2 \alpha^2 \sum_{t=0}^T{1}
                \\
                & \le \frac{{\left( C_i^{(0)} \right)}^2}{1 - (1 - \beta_i)^2} + \frac{2 \gamma_i B_i \alpha C_i^{(0)}}{1 - (1 - \beta_i)} + \gamma_i^2 B_i^2 \alpha^2 (T+1).
            \end{aligned}
        \end{equation}
        Now, taking the average of the above expression, we obtain
        \begin{equation}
            \begin{aligned}
                & \sum_{t=0}^T{\overline{{(C^{(t)})}^2}} = \frac1m \sum_{i=1}^m{\left( \frac{{\left( C_i^{(0)} \right)}^2}{1 - (1 - \beta_i)^2} + \frac{2 \gamma_i B_i \alpha C_i^{(0)}}{1 - (1 - \beta_i)} + \gamma_i^2 B_i^2 \alpha^2 (T+1) \right)}
                \\
                & \le \frac{\sum_{i=1}^m{{\left( C_i^{(0)} \right)}^2}}{m \left( 1 - {(1 - \mini{\beta_i})}^2 \right)} + \frac{2 \maxi{\gamma_i B_i} \alpha}{\mini{\beta_i} m} \sum_{i=1}^m{C_i^{(0)}} + \maxi{\gamma_i B_i}^2 \alpha^2 (T+1).
            \end{aligned}
        \end{equation}
    \end{proof}
\end{lemma}

\begin{theorem}[Stationarity point for constant learning rate and adaptive DP] \label{theorem:adaptive}
    Let Assumptions~\ref{assump:smooth}-\ref{assump:bound} and \ref{assump:clipnum} hold. Also, let a constant learning rate $\alpha_i^{(t)} = \alpha$ be adopted for all clients $i \in \mathcal{M}$, and an adaptive DP algorithm be used, i.e., adaptive clipping threshold $C_i^{(t+1)} = (1 - \beta_i) C_i^{(t)} + \beta_i \gamma_i \| \Delta \theta_i^{(t)} \|$ and an adaptive noise variance given as $(\sigma_{0,i}^2)^{(t)} = 2 z_{\Delta,i}^2 (1 - \beta_i)^2 (C_i^{(t)})^2$ and $(\sigma_{1,i}^2)^{(t)} = 2 z_{\Delta,i}^2 \beta_i^2 \gamma_i^2$. For simplicity, let all clients use the same moving average rate $\beta_i = \beta$, same quantile $\gamma_i = \gamma$ and same noise multiplier $z_{\Delta,i} = z_\Delta$. Then, Proposition~\ref{prop:finalloss_adaptive} and Lemma~\ref{lemma:recursive} imply that if the learning rate $\alpha$ satisfies
    \begin{equation}
        \alpha < \frac{2{(1 - p)}^m}{3 \bar{L} (\bar{\rho}_1^2 + \bar{\zeta}_1^2 + 1) \Gamma_2} \frac1{1 + 2 z_\Delta^2 \beta^2 \gamma^2},
    \end{equation}
    we will have
    \begin{equation} \label{eqn:adaptive}
        \begin{aligned}
            \frac{\sum_{t=0}^T{\norm{\nabla{F}(\theta^{(t)})}^2}}{T+1} \le & \,\, \frac{F(\theta^{(0)}) - F^\star}{\omega \alpha (T+1)}
            \\
            & + \frac{\bar{L}}{2 \omega} \left( 1 + z_\Delta^2 (1 - \beta)^2 \right) \Bigg[ \frac{{\left( C^{(0)} \right)}^2}{\left( 1 - (1 - \beta)^2 \right) \alpha (T+1)} + \frac{2 \gamma B C^{(0)}}{\beta (T+1)} + \gamma^2 B^2 \alpha \Bigg]
            \\
            & + \frac{3 \bar{L}}{2\omega} (\bar{\rho}_0^2 + \bar{\zeta}_0^2) \alpha \left( 1 + 2 z_\Delta^2 \beta^2 \gamma^2 \right).
        \end{aligned}
    \end{equation}
    where $\omega = {(1 - \maxi{p_i})}^m \left( 1 - \frac1{\Gamma} \right)$.

    \begin{proof}
        For the noise variance of Eq.~\eqref{eqn:noiseupdate}, it follows that
        \begin{equation}
            ( \overline{\sigma_0^2} )^{(t)} = \frac1m \sum_{i=1}^m{( \sigma_{0,i}^2 )^{(t)}} \le \frac1m \sum_{i=1}^m{2 z_{\Delta,i}^2 (1 - \beta_i)^2 {\left( C_i^{(t)} \right)}^2} \le \maxi{z_{\Delta,i} (1 - \beta_i)}^2 {\left( C^{(t)} \right)}^2.
        \end{equation}
        As a result, we will obtain using Lemma~\ref{lemma:recursive} that
        \begin{equation}
            \begin{aligned}
                & \sum_{t=0}^T{\left( \overline{{(C^{(t)})}^2} + ( \overline{\sigma_0^2} )^{(t)} \right)} \le \left( 1 + \maxi{z_{\Delta,i} (1 - \beta_i)}^2 \right) \sum_{t=0}^T{\overline{{(C^{(t)})}^2}}
                \\
                & \begin{aligned}
                    \,\, \le \left( 1 + \maxi{z_{\Delta,i} (1 - \beta_i)}^2 \right) \Bigg[ \frac{{\left( C^{(0)} \right)}^2}{1 - (1 - \mini{\beta_i})^2} & + \frac{2 \maxi{\gamma_i B_i} \alpha \bar{C}^{(0)}}{\mini{\beta_i}}
                    \\
                    & + \maxi{\gamma_i B_i}^2 \alpha^2 (T+1) \Bigg]
                \end{aligned}
            \end{aligned}
        \end{equation}
        Plugging this value in Proposition~\ref{prop:finalloss_adaptive} gives us
        \begin{equation}
            \begin{aligned}
                \bbE{F(\theta^{(T+1)})} \le & \,\, F(\theta^{(0)}) - \omega \alpha \sum_{t=0}^T{\norm{\nabla{F}(\theta^{(t)})}^2}
                \\
                & \begin{aligned}
                    + \frac12 \bar{L} & \left( 1 + \maxi{z_{\Delta,i} (1 - \beta_i)}^2 \right) \Bigg[ \frac{{\left( C^{(0)} \right)}^2}{1 - (1 - \mini{\beta_i})^2}
                    \\
                    & + \frac{2 \maxi{\gamma_i B_i} \alpha C^{(0)}}{\mini{\beta_i}} + \maxi{\gamma_i B_i}^2 \alpha^2 (T+1) \Bigg]
                \end{aligned}
                \\
                & + \frac32 \bar{L} (\bar{\rho}_0^2 + \bar{\zeta}_0^2) \alpha^2 \left( 1 + \maxi{\sigma_{1,i}^2} \right) (T+1).
            \end{aligned}
        \end{equation}
        Rearranging this inequality yields
        \begin{equation}
            \begin{aligned}
                \frac{\sum_{t=0}^T{\norm{\nabla{F}(\theta^{(t)})}^2}}{T+1} \le & \,\, \frac{F(\theta^{(0)}) - F^\star}{\omega \alpha (T+1)}
                \\
                & \begin{aligned}
                    + \frac{\bar{L}}{2 \omega} & \left( 1 + \maxi{z_{\Delta,i} (1 - \beta_i)}^2 \right) \Bigg[ \frac{{\left( C^{(0)} \right)}^2}{\left( 1 - (1 - \mini{\beta_i})^2 \right) \alpha (T+1)}
                    \\
                    & + \frac{2 \maxi{\gamma_i B_i} C^{(0)}}{\mini{\beta_i} (T+1)} + \maxi{\gamma_i B_i}^2 \alpha \Bigg]
                \end{aligned}
                \\
                & + \frac{3 \bar{L}}{2\omega} (\bar{\rho}_0^2 + \bar{\zeta}_0^2) \alpha \left( 1 + \maxi{\sigma_{1,i}^2} \right).
            \end{aligned}
        \end{equation}
    \end{proof}
\end{theorem}

In Eq.~\eqref{eqn:adaptive}, we show that when using a constant learning rate $\alpha$ with adaptive DP, the convergence rate of our FL algorithm achieves the same rate $\mathcal{O}(1/T)$ as centralized training to a stationarity neighborhood of $\mathcal{O}(\frac{\bar{L}}{2 \omega} \left( 1 + z_\Delta^2 (1 - \beta)^2 \right) \gamma^2 B^2 \alpha)$. We observe that compared to Eq.~\eqref{eqn:fixed}, the stationarity gap only depends on the learning rate $\alpha$ since the clipping threshold is being adjusted dynamically. Similar to Theorem~\ref{theorem:fixed}, setting $\alpha \propto 1/\sqrt{T}$ will result in a slower convergence rate of $\mathcal{O}(1 / \sqrt{T})$ but the stationarity gap will be zero.

\end{document}